%% file: HumanDynamicsModelling.tex
\documentclass[10pt,twocolumn,letterpaper]{article}

\pdfoutput=1
\usepackage{iccv}
\usepackage{times}
\usepackage{epsfig}
\usepackage{graphicx}
\usepackage{amsmath}
\usepackage{amssymb}

\usepackage{titling}
\usepackage{float}
\usepackage{multirow}
\usepackage[pagebackref=true,breaklinks=true,letterpaper=true,colorlinks,bookmarks=false]{hyperref}

\newcommand{\comment}[1]{}

\newcommand{\NEW}[1]{\textcolor{cyan}{#1}}

\iccvfinalcopy 

\def\iccvPaperID{3937} 

\ificcvfinal\fi
\begin{document}
\title{Learning Trajectory Dependencies for Human Motion Prediction}
\date{\vspace{-0.5cm}}
\author{Wei Mao$^1$, \;\;Miaomiao Liu$^{1,3}$,\;\; Mathieu Salzmann$^2$,\;\; Hongdong Li$^{1,3}$\\
$^1$Australian National University, $^2$CVLab, EPFL,$^3$Australia Centre for Robotic Vision\\
{\tt\small \{wei.mao, miaomiao.liu, hongdong.li\}@anu.edu.au,}\;\;{\tt\small mathieu.salzmann@epfl.ch}
}
\maketitle
\thispagestyle{empty}
\input{0_abstract.tex}
\input{1_introduction.tex}
\input{2_relatedwork.tex}
\input{3_approach.tex}
\input{4_experiments.tex}
\input{5_conclusion.tex}
\newpage
\setcounter{section}{0}
\setcounter{figure}{0}
\setcounter{table}{0}
\input{HumanDynamicsSupplementary.tex}

{\small
\bibliographystyle{ieee_fullname}
\bibliography{HumanDynamicsModelling}
}

\end{document}

%% file: 0_abstract.tex
\begin{abstract}
Human motion prediction, \ie,~forecasting future body poses given observed pose sequence, has typically been tackled with recurrent neural networks (RNNs).~However, as evidenced by prior work,~the resulted RNN models suffer from prediction errors accumulation, leading to undesired discontinuities in motion prediction.~In this paper, we propose a simple feed-forward deep network for motion prediction, which takes into account both temporal smoothness and spatial dependencies among human body joints. In this context, we then propose to encode temporal information by working in trajectory space, instead of the traditionally-used pose space.~This alleviates us from manually defining the range of temporal dependencies (or temporal convolutional filter size, as done in previous work).~Moreover, spatial dependency of human pose is encoded by treating a human pose as a generic graph (rather than a human skeletal kinematic tree) formed by links between every pair of body joints.~Instead of using a pre-defined graph structure, we design a new graph convolutional network to learn graph connectivity automatically. This allows the network to capture long range dependencies beyond that of human kinematic tree.~We evaluate our approach on several standard benchmark datasets for motion prediction, including Human3.6M, the CMU motion capture dataset and 3DPW.~Our experiments clearly demonstrate that the proposed approach achieves state of the art performance, and is applicable to both angle-based and position-based pose representations. The code is available at \url{https://github.com/wei-mao-2019/LearnTrajDep}
\end{abstract}

%% file: 1_introduction.tex
\section{Introduction}\label{intro}
Human motion prediction is key to the success of applications where one needs to forecast the future, such as human robot interaction~\cite{koppula2013anticipating}, autonomous driving~\cite{paden2016survey} and human tracking~\cite{gong2011multi}. 
While traditional data-driven approaches, such as Hidden Markov Model~\cite{brand2000style} and Gaussian Process latent variable models~\cite{wang2008gaussian}, have proved effective for simple periodic motions and acyclic motions, such as walking and golf swing, more complicated ones are typically tackled using deep networks~\cite{fragkiadaki2015recurrent,JainZSS16,Butepage_2017_CVPR,Martinez_2017_CVPR,gui2018adversarial,LiZLL18}.

\begin{figure}
 \centering
   \includegraphics[width=1\linewidth]{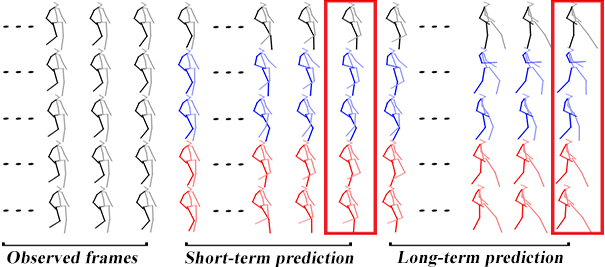}
    \caption{{\bf Human motion prediction.} The left frames correspond to the observations. From top to bottom, we show the ground truth, and predictions obtained by the methods of~\cite{Martinez_2017_CVPR} and~\cite{LiZLL18}, and by our approach on joint angles and 3d coordinates. Our predictions better match the ground truth.
    }
    \vspace{-0.4cm}
    \label{fig:human-motion-prediction-task}
\end{figure}

Because of the temporal nature of the signal of interest, the most common trend consists of using Recurrent Neural Networks (RNNs)~\cite{fragkiadaki2015recurrent,JainZSS16,Martinez_2017_CVPR,gui2018adversarial}. However, as argued in~\cite{gui2018adversarial,LiZLL18}
, besides their well-known training difficulty~\cite{pascanu2013difficulty}, RNNs for motion prediction suffer from several drawbacks: First, existing works~\cite{fragkiadaki2015recurrent,Martinez_2017_CVPR} that use the estimation at the current RNN step as input to the next prediction
 tend to accumulate errors throughout the generated sequence, leading to unrealistic predictions at inference time.~Second, as observed in~\cite{LiZLL18,Martinez_2017_CVPR}, earlier RNN-based methods~\cite{fragkiadaki2015recurrent,JainZSS16} 
often produce strong discontinuities between the last observed frame and the first predicted one. These discontinuities are partially due to the frame-by-frame regression procedure that does not encourage global smoothness of the sequence~\cite{gui2018adversarial}. As a consequence, several works have proposed to rely on feed-forward networks for motion prediction~\cite{Butepage_2017_CVPR,LiZLL18}. In this paper, we introduce a new feed-forward approach to motion prediction, leading to more accurate predictions than RNN ones, as illustrated in Fig.~\ref{fig:human-motion-prediction-task}.

When using feed-forward networks for a time-related problem such as motion prediction, the question of how to encode the temporal information naturally arises. In~\cite{Butepage_2017_CVPR,LiZLL18}, this was achieved by using convolutions across time on the observed poses. The temporal dependencies that such an approach can encode, however, strongly depend on the size of the convolutional filters. 

To remove such a dependency, here, we introduce a drastically different approach to modeling temporal information for motion prediction. Inspired by ideas from the nonrigid structure-from-motion literature~\cite{akhter2009nonrigid}, we propose to represent human motion in trajectory space instead of pose space, and thus adopt the Discrete Cosine Transform (DCT) to encode temporal information.
Specifically, we represent the temporal variation of each human joint as a linear combination of DCT bases, and, given the DCT coefficients of the observed poses, learn to predict those of the future ones. This strategy applies to both angle-based pose representations and 3D joint positions. As discussed in our experiments, the latter has the advantage of not suffering from ambiguities, in contrast to angle-based ones, where two different sets of angles can represent the exact same pose. As a consequence, reasoning in terms of 3D joint positions allows one not to penalize configurations that differ from ground truth while depicting equivalent poses.

The other question that arises when working with human pose is how to encode the spatial dependencies among the joints. In~\cite{Butepage_2017_CVPR}, this was achieved by exploiting the human skeleton, and in~\cite{LiZLL18} by defining a relatively large spatial filter size. While the former does not allow one to model dependencies across different limbs, such as left-right symmetries, the latter again depends on the size of the filters.

In this paper, we propose to overcome these two issues by exploiting graph convolutions~\cite{kipf2016semi}. However, instead of using a pre-defined, sparse graph as in~\cite{kipf2016semi}, we introduce an approach to learning the graph connectivity. This strategy allows the network to capture joint dependencies that are neither restricted to the kinematic tree, nor arbitrarily defined by a convolutional kernel size.

In summary, our contributions are (i) a natural way to encode temporal information in feed-forward networks for motion prediction via the DCT; (ii) learnable graph convolutional networks to capture the spatial structure of the motion data. Our experiments on standard human motion prediction benchmarks evidence the benefits of our approach; our model yields state-of-the-art results in all cases. 

%% file: 2_relatedwork.tex
\section{Related Work}
\noindent{\bf RNN-based human motion prediction.}
Because of their success at sequence-to-sequence prediction~\cite{sutskever2011generating,kiros2015skip}, RNNs have become the {\it de facto} model for human motion prediction~\cite{fragkiadaki2015recurrent,JainZSS16,Martinez_2017_CVPR}. This trend was initiated by Fragkiadaki \etal~\cite{fragkiadaki2015recurrent}, who proposed an~\emph{Encoder-Recurrent-Decoder~(ERD)} model that incorporates a nonlinear encoder and decoder before and after recurrent layers. Error accumulation was already observed in this work, and a~\emph{curriculum learning} strategy was adopted during training to prevent it. 
In~\cite{JainZSS16}, Jain \etal proposed to further encode the spatial and temporal structure of the pose prediction problem via a~\emph{Structural-RNN} model relying on high-level spatio-temporal graphs. These graphs, however, were manually designed, which limits the flexibility of the framework, not letting it discover long-range interactions between different limbs. While the two previous methods directly estimated absolute human poses, Martinez \etal~\cite{Martinez_2017_CVPR} introduced a~\emph{residual} architecture to predict velocities. Interestingly, it was shown in this work that a simple zero-velocity baseline, i.e., constantly predicting the last observed pose, led to better performance than~\cite{fragkiadaki2015recurrent,JainZSS16}. While~\cite{Martinez_2017_CVPR} outperformed this baseline, the predictions produced by the RNN still suffer from discontinuities between the observed poses and the predicted future ones. To overcome this, Gui \etal proposed to rely on adversarial training, so as to generate smooth sequences that are indistinguishable from real ones~\cite{gui2018adversarial}. While this approach constitutes the state of the art, its use of an adversarial classifier, which notoriously complicates training~\cite{ArjovskyB17}, makes it difficult to deploy on new datasets.\\
\noindent{\bf Feed-forward approaches to human motion prediction.} Feed-forward networks, such as fully-connected and convolutional ones, were studied as an alternative solution to avoiding the discontinuities produced by RNNs~\cite{Butepage_2017_CVPR,LiZLL18}. In particular, in~\cite{Butepage_2017_CVPR}, Butepage \etal proposed to treat a recent pose history as input to a fully-connected network, and introduced different strategies to encode additional temporal information via convolutions and spatial structure by exploiting the kinematic tree. 
The use of a kinematic tree, however, does not reflect the fact that, as discussed in~\cite{LiZLL18}, stable motion requires synchronizing different body parts, even distant ones not directly connected by the kinematic tree. To capture such dependencies, Li \etal~\cite{LiZLL18} built a convolutional sequence-to-sequence model processing a 2 dimensional matrix whose columns represent the pose at every time step. The range of the spatial and temporal dependencies captured by this model is then determined by the size of the convolutional filters. In this paper, as in~\cite{Butepage_2017_CVPR,LiZLL18}, we also rely on a feed-forward network for motion prediction. However, we introduce a drastically different way to modeling temporal information, which, in contrast to~\cite{Butepage_2017_CVPR,LiZLL18}, does not require manually defining convolutional kernel sizes. Specifically, we propose to perform motion prediction in trajectory space instead of pose space. Furthermore, to model the spatial dependencies between the joints, we propose to exploit graph convolutional networks.
\begin{figure*}[!ht]
    \centering
      \includegraphics[width=0.9\textwidth]{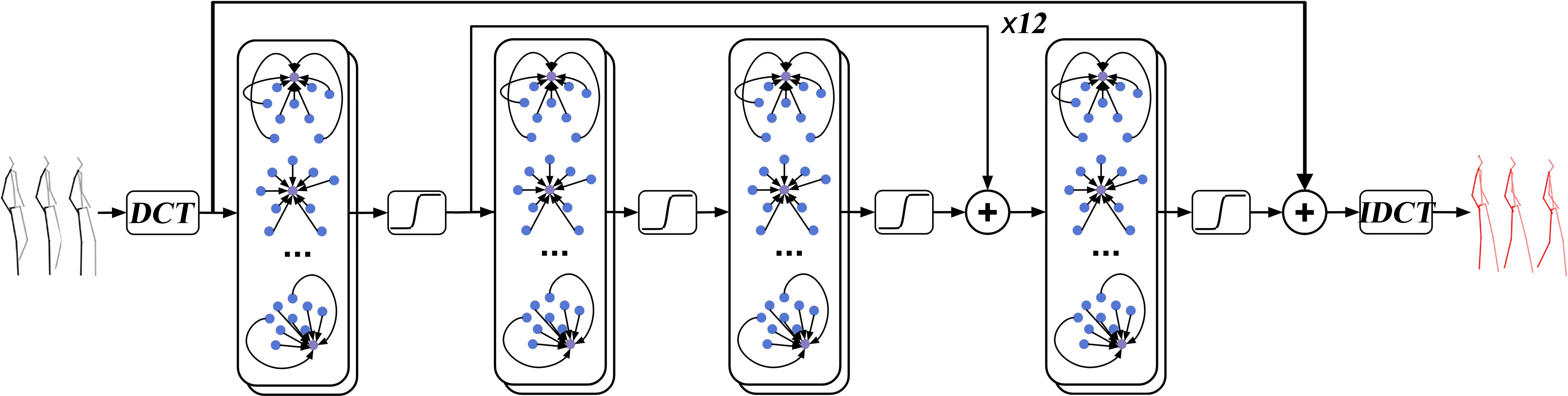}
      \caption{{\bf Network architecture.} We first apply the DCT to encode temporal pose information in trajectory space. The DCT coefficients are treated as features input to graph convolutional layers. We use 12 blocks of graph convolutional layers with residual connections and two additional graph convolutional layers, one at the beginning and one at the end, to encode the temporal information and decode the features to the residual DCT coefficients, respectively. In each block, we depict how our framework aggregates information from multiple nodes via learned adjacency matrices.}
      \vspace{-0.5cm}
      \label{fig:net-structure}
\end{figure*}
\noindent{\bf Graph Convolutional Networks (GCNs).} GCNs generalize the convolution operation to data whose structure is defined by a graph, such as user data from social networks, data defined on 3D meshes and gene data on biological regulatory networks~\cite{BrunaZSL13,defferrard2016convolutional}. The main advances in this context can be categorized as spectral~\cite{kipf2016semi} and non-spectral~\cite{velivckovic2017graph} methods. In particular, Kipf and Welling~\cite{kipf2016semi} use  filters that depend on the graph structure, which limits the generality of their approach. By contrast, Veli\u ckovi\'c \etal~\cite{velivckovic2017graph} rely on self-attention to determine the neighborhood structure to be considered, thus providing more flexibility to the network. A straightforward approach to exploiting graph convolutions for motion prediction would consist of relying on the kinematic tree to define the graph. This strategy has been employed for action recognition~\cite{yan2018spatial}, by using a GCN to capture the temporal and spatial dependencies of human joints via a graph defined on temporally connected kinematic trees.~For motion prediction, however, this would suffer from the same limitations as the strategy of~\cite{Butepage_2017_CVPR} discussed above. Therefore, here, inspired by~\cite{velivckovic2017graph}, we design a GCN able to adaptively learn the necessary connectivity for the motion prediction task at hand. 

%% file: 3_approach.tex
\section{Our Approach}
Let us now introduce our approach to human motion prediction.
As existing methods, we assume to be given a history motion sequence ${\bf X}_{1:N} = [{\bf x}_1,{\bf x}_2,{\bf x}_3,\cdots,{\bf x}_N]$ consisting of $N$ consecutive human poses, where ${\bf x}_i\in \mathbb{R}^K$, with $K$ the number of parameters describing each pose. Our goal then is to predict the poses ${\bf X}_{N+1:N+T}$ for the future $T$ time steps. To this end, we propose to make use of a feed-forward deep network that models the temporal and spatial structure of the data. Below, we introduce our approach to encoding these two types of information and then provide the details of our network architecture.

\subsection{\emph{DCT}-based Temporal Encoding}\label{subsec:tmpencoding}
In the motion prediction literature, the two standard ways to represent human pose are joint angles and 3D joint coordinates. These two representations, however, are purely static. Here, instead, we propose to directly encode the temporal nature of human motion in our representation and work in trajectory space. Note that, ultimately, we nonetheless need to produce human poses in a standard representation, and, as evidenced by our experiments, our formalism applies to both of the above-mentioned ones.

Our temporal encoding aims to capture the motion pattern of each joint. Recall that each column of ${\bf X}_{1:N}$ represents the human pose at a specific time step. Conversely, each row of ${\bf X}_{1:N}$ describes the motion of each joint (angle or coordinate). Let us denote by $\Tilde{{\bf x}}_k=( x_{k,1}, x_{k,2}, x_{k,3},\cdots, x_{k,N})$ the trajectory for the $k^{th}$ joint across $N$ frames. While one could directly use such trajectories as input and output for motion prediction,
inspired by ideas from the nonrigid-structure-from-motion literature~\cite{akhter2009nonrigid}, we propose to adopt a trajectory representation based on the Discrete Cosine Transform (DCT). The main motivation behind this is that, by discarding the high frequencies, the DCT can provide a more compact representation, which nicely captures the smoothness of human motion, particularly in terms of 3D coordinates. Detailed analysis about the number of DCT coefficients used is in the supplementary material.

Specifically, given a trajectory $\Tilde{{\bf x}}_k$, the corresponding $l^{th}$ DCT coefficient can be computed as
\vspace{-0.2cm}
\begin{equation}
    \resizebox{0.9\linewidth}{!}{$C_{k,l} = \sqrt{\frac{2}{N}}\sum_{n=1}^{N}x_{k,n}\frac{1}{\sqrt{1+\delta_{l1}}}\cos\left(\frac{\pi}{2N}(2n-1)(l-1)\right)\;,$}
    \label{eq:dct}
    \vspace{-0.2cm}
\end{equation}
where $\delta_{ij}$ denotes the~\emph{Kronecker} delta function with 
\vspace{-0.2cm}
\begin{equation}
  \resizebox{0.3\linewidth}{!}{$\delta_{ij} = \begin{cases}
  1 & \text{if}\ i=j\\
  0 & \text{if}\ i\neq j.
  \end{cases}$}
  \vspace{-0.2cm}
\end{equation}
In practice, $l\in\{1,2,\cdots,N\}$, but one can often ignore the higher values, which, in our context, translates to removing the high motion frequencies. In short, Eq.~\ref{eq:dct} allows us to model the temporal information of each joint using DCT coefficients.
Given such coefficients, the original pose representation (angles or coordinates) can be obtained via the Inverse Discrete Cosine Transform (IDCT) as
\vspace{-0.2cm}
\begin{equation}
    \resizebox{0.9\linewidth}{!}{$x_{k,n} =\sqrt{\frac{2}{N}}\sum_{l=1}^{N}C_{k,l}\frac{1}{\sqrt{1+\delta_{l1}}}\cos\left(\frac{\pi}{2N}(2n-1)(l-1)\right)\;,$}
    \vspace{-0.2cm}
\end{equation}
where $n\in\{1,2,\cdots,N\}$. Note that, if all DCT coefficients are used, the resulting representation is lossless. However, as mentioned before, truncating some of the high frequencies can prevent generating jittery motion.

To make use of the DCT representation, instead of treating motion prediction as the problem of learning a mapping from ${\bf X}_{1:N}$ to ${\bf X}_{N+1:N+T}$, we reformulate it as one of learning a mapping between observed and future DCT coefficients. Specifically, given a temporal sequence ${\bf X}_{1:N}$, we first replicate the last pose, ${\bf x}_N$, $T$ times to generate a temporal sequence of length $N+T$. We then compute the DCT coefficients of this sequence, and aim to predict those of the true future sequence ${\bf X}_{1:N+T}$. This naturally translates to estimating a residual vector in frequency space and was motivated by the zero-velocity baseline in~\cite{Martinez_2017_CVPR}. As will be shown in our experiments, this residual approach, with padding by replicating the last pose, has proven much more effective than other strategies.

Our DCT representations could be directly employed in a standard fully-connected network, either by stacking the DCT representations of all joints in a single vector, which would yield to a network with many parameters, or by treating the different DCT coefficients as different channels, thus using a $K \times L$ matrix as input to the network, with $L$ the number of retained DCT coefficients. While this latter strategy results in a more compact network, it does not model the spatial dependencies between the joints. In the next section, we introduce an approach to doing so using GCNs.

\subsection{Graph Convolutional Layer}
To encode the spatial structure of human pose, we make use of GCNs~\cite{kipf2016semi,velivckovic2017graph}. Here, instead of relying on a predefined, sparse graph, as in~\cite{kipf2016semi}, we propose to learn the graph connectivity during training, thus essentially learning the dependencies between the different joint trajectories.

To this end, let us assume that the human body is modeled as a fully-connected graph with $K$ nodes. The strength of the edges in this graph can then be represented by a weighted adjacency matrix ${\bf A}\in \mathbb{R}^{K \times K}$. A graph convolutional layer $p$ then takes as input a matrix ${\bf H}^{(p)}\in \mathbb{R}^{K\times F}$, with $F$ the number of features output by the previous layer. For example, for the first layer, the network takes as input the $K \times L$\comment{\footnote{\NEW{We use all DCT coefficients for motion prediction in angle space while in 3D coordinate space, partial coefficients are used (see Supp. Mat.).}}} matrix of DCT coefficients. Given this information and a set of trainable weights ${\bf W}^{(p)}\in \mathbb{R}^{F \times \hat{F}}$, a graph convolutional layer outputs a matrix of the form
\vspace{-0.2cm}
\begin{equation}
    \resizebox{0.5\linewidth}{!}{${\bf H}^{(p+1)} = \sigma({\bf A}^{(p)}{\bf H}^{(p)}{\bf W}^{(p)})\;,$}
    \vspace{-0.2cm}
\end{equation}
where ${\bf A}^{(p)}$ is the trainable weighted adjacency matrix for layer $p$ and $\sigma(\cdot)$ is an activation function, such as $tanh(\cdot)$.

Following the standard deep learning formalism, multiple such layers can be stacked to form a GCN. Since all operations are differentiable, w.r.t. both ${\bf A}^{(p)}$ and ${\bf W}^{(p)}$, the resulting network can be trained using standard back-propagation. In the next section, we provide additional detail about the network structure used in our experiments.

\subsection{Network Structure}
As discussed in Section~\ref{subsec:tmpencoding}, we aim to learn the residuals between the input and output DCT representations. More precisely, we learn the residuals between the DCT coefficients obtained from the input sequence with replicated last pose, and that of the sequence ${\bf X}_{1:N+T}$. We therefore design a residual graph convolutional network. The network structure is shown in Fig.~\ref{fig:net-structure}. It consists of 12 residual blocks, each of which comprises 2 graph convolutional layers and two additional graph convolutional layers, one at the beginning and one at the end, to encode the temporal information and decode the features to the residual DCT coefficients, respectively.  Each layer $p$ relies on a learnable weight matrix ${\bf W}^{(p)}$ of size $256\times 256$ and a learnable weighted adjacency matrix ${\bf A}^{(p)}$. Using a different learnable ${\bf A}$ for every graph convolutional layer allows the network to adapt the connectivity for different operations. This gives our framework a greater capacity than a GCN with a fixed adjacency matrix. Nevertheless, because, in each layer $p$, the weight matrix ${\bf W}^{(p)}$ is shared by the different joints to further extract motion patterns from feature matrix, the overall network remains compact; the size of the models used in our experiments is around $2.6M$ for both angle and 3D representations.

\subsection{Training}
As mentioned before, joint angles and 3D coordinates are the two standard representations for human pose, and we will evaluate our approach on both. Below, we discuss the loss function we use to train our network in each case.
For joint angles, following the literature, we use an exponential map representation. Given the training angles, we apply the DCT to obtain the corresponding coefficients, train our model and employ the IDCT to the predicted DCT coefficients so as to retrieve the corresponding angles ${\bf X}_{1:N+T}$. To train our network, we use the average $\ell_1$ distance between the ground-truth joint angles and the predicted ones. Formally, for one training sample, this gives the loss
\vspace{-0.2cm}
\begin{equation}
    \resizebox{0.7\linewidth}{!}{$\ell_{a} = \frac{1}{(N+T)K}\sum_{n=1}^{N+T}\sum_{k=1}^K|\hat{x}_{k,n}-x_{k,n}|\;,$}
    \vspace{-0.2cm}
\end{equation}
\normalsize
where $\hat{x}_{k,n}$ is the predicted $k^{th}$ angle in frame $n$ and $x_{k,n}$ the corresponding ground-truth one. Note that we sum $\ell_1$ errors over both the future \emph{and} observed time steps. This provides us with additional signal to learn to predict the DCT coefficients, which represent the \emph{entire} sequence.

For the coordinate-based representation, we adopt the standard body model of~\cite{h36m_pami} to convert the joint angles to 3D coordinates. The 3D joint positions are then pre-processed so as to be centred at the origin, and the global rotations are removed. Going from 3D coordinates to DCT coefficients and back follows exactly the same procedure as in the angle case.
To train our model, we then make use of the Mean Per Joint Position Error (MPJPE) proposed in~\cite{h36m_pami}, which, for one training sample, translates to the loss
\vspace{-0.2cm}
\begin{equation}
    \resizebox{0.7\linewidth}{!}{$ \ell_m = \frac{1}{J(N+T)}\sum_{n=1}^{N+T}\sum_{j=1}^{J}\|\hat{\textbf{p}}_{j,n}-\textbf{p}_{j,n}\|^2\;,$}
    \vspace{-0.2cm}
\end{equation}
where $\hat{\textbf{p}}_{j,n}\in \mathbb{R}^{3}$ denotes the predicted $j$th joint position in frame $n$, ${\bf p}_{j,n}$ the corresponding ground-truth one, and $J$ the number of joints in the human skeleton.

\begin{table*}[ht]
\centering
\resizebox{0.9\textwidth}{!}{%
\begin{tabular}{ccccc|cccc|cccc|cccc}
  & \multicolumn{4}{c}{Walking} & \multicolumn{4}{c}{Eating} & \multicolumn{4}{c}{Smoking} & \multicolumn{4}{c}{Discussion} \\
  milliseconds& 80 & 160 & 320 & 400 & 80 & 160 & 320 & 400 & 80 & 160 & 320 & 400 & 80 & 160 & 320 & 400 \\ \hline
  zero-velocity \cite{Martinez_2017_CVPR} & 0.39 & 0.68 & 0.99 & 1.15 & 0.27 & 0.48 & 0.73 & 0.86 & 0.26 & 0.48 & 0.97 & 0.95 & 0.31 & 0.67 & 0.94 & 1.04 \\
 Residual sup. \cite{Martinez_2017_CVPR} & 0.28 & 0.49 & 0.72 & 0.81 & 0.23 & 0.39 & 0.62 & 0.76 & 0.33 & 0.61 & 1.05 & 1.15 & 0.31 & 0.68 & 1.01 & 1.09\\
 convSeq2Seq \cite{LiZLL18} & 0.33 & 0.54 & 0.68 & 0.73 & 0.22 & 0.36 & 0.58 & 0.71 & 0.26 & 0.49 & 0.96 & 0.92 & 0.32 & 0.67 & 0.94 & 1.01\\
 AGED w/o adv \cite{gui2018adversarial} & 0.28 & 0.42 & 0.66 & 0.73 & 0.22 & 0.35 & 0.61 & 0.74 & 0.3 & 0.55 & 0.98 & 0.99 & 0.30 & 0.63 & 0.97 & 1.06 \\
 AGED w/adv \cite{gui2018adversarial} & 0.22 & 0.36 & 0.55 & 0.67 & 0.17 & \textbf{0.28} & 0.51 & 0.64 & 0.27 & 0.43 & \textbf{0.82} & 0.84 & 0.27 & 0.56 & \textbf{0.76} & \textbf{0.83}\\ \hline
 ours & \textbf{0.18} & \textbf{0.31} & \textbf{0.49} & \textbf{0.56} & \textbf{0.16} & 0.29 & \textbf{0.50} & \textbf{0.62} & \textbf{0.22} & \textbf{0.41} & 0.86 & \textbf{0.80} & \textbf{0.20} & \textbf{0.51} & 0.77 & 0.85\\\hline
\end{tabular}
}
\resizebox{\textwidth}{!}{%
\begin{tabular}{ccccc|cccc|cccc|cccc|cccc|cccc}
  & \multicolumn{4}{c}{Directions} & \multicolumn{4}{c}{Greeting} & \multicolumn{4}{c}{Phoning} & \multicolumn{4}{c}{Posing}&\multicolumn{4}{c}{Purchases}&\multicolumn{4}{c}{Sitting} \\
  milliseconds& 80 & 160 & 320 & 400 & 80 & 160 & 320 & 400 & 80 & 160 & 320 & 400 & 80 & 160 & 320 & 400& 80 & 160 & 320 & 400& 80 & 160 & 320 & 400 \\ \hline
  zero-velocity \cite{Martinez_2017_CVPR} & 0.39 & 0.59 & 0.79 & 0.89 & 0.54 & 0.89 & 1.30 & 1.49 & 0.64 & 1.21 & 1.65 & 1.83 & 0.28 & 0.57 & 1.13 & 1.37 & 0.62 & 0.88 & 1.19 & 1.27 & 0.40 & 1.63 & 1.02 & 1.18\\
  Residual sup. \cite{Martinez_2017_CVPR} & 0.26 & 0.47 & 0.72 & 0.84 & 0.75 & 1.17 & 1.74 & 1.83 & 0.23 & 0.43 & 0.69 & 0.82 & 0.36 & 0.71 & 1.22 & 1.48 & 0.51 & 0.97 & 1.07 & 1.16 & 0.41 & 1.05 & 1.49 & 1.63\\
  convSeq2Seq \cite{LiZLL18} & 0.39 & 0.60 & 0.80 & 0.91 & 0.51 & 0.82 & 1.21 & 1.38 & 0.59 & 1.13 & 1.51 & 1.65 & 0.29 & 0.60 & 1.12 & 1.37 & 0.63 & 0.91 & 1.19 & 1.29 & 0.39 & 0.61 & 1.02 &1.18\\
  AGED w/o adv \cite{gui2018adversarial}& 0.26 & 0.46 & 0.71 & 0.81 & 0.61 & 0.95 & 1.44 & 1.61 & 0.23 & 0.42 & 0.61 & 0.79 & 0.34 & 0.70 & 1.19 & 1.40 & 0.46 & 0.89 & 1.06 & 1.11 & 0.46 & 0.87 & 1.23 & 1.51 \\
  AGED w/adv \cite{gui2018adversarial} & \textbf{0.23} & \textbf{0.39} & \textbf{0.63} & \textbf{0.69} & 0.56 & 0.81 & 1.30 & 1.46 & \textbf{0.19} & \textbf{0.34} & \textbf{0.50} & \textbf{0.68} & 0.31 & 0.58 & 1.12 & 1.34 & 0.46 & 0.78 & \textbf{1.01} & \textbf{1.07} & 0.41 & 0.76 & 1.05 & 1.19\\\hline
  Ours & 0.26 & 0.45 & 0.71 & 0.79 & \textbf{0.36} & \textbf{0.60} & \textbf{0.95} & \textbf{1.13} & 0.53 & 1.02 & 1.35 & 1.48 & \textbf{0.19} & \textbf{0.44} & \textbf{1.01} & \textbf{1.24} & \textbf{0.43} & \textbf{0.65} & 1.05 & 1.13 & \textbf{0.29} & \textbf{0.45} & \textbf{0.80} & \textbf{0.97}\\ \hline
  & \multicolumn{4}{c}{Sitting Down} & \multicolumn{4}{c}{Taking Photo} & \multicolumn{4}{c}{Waiting} & \multicolumn{4}{c}{Walking Dog}&\multicolumn{4}{c}{Walking Together}&\multicolumn{4}{c}{Average} \\
  milliseconds& 80 & 160 & 320 & 400 & 80 & 160 & 320 & 400 & 80 & 160 & 320 & 400 & 80 & 160 & 320 & 400& 80 & 160 & 320 & 400& 80 & 160 & 320 & 400 \\ \hline
  zero-velocity \cite{Martinez_2017_CVPR} & 0.39 & 0.74 & 1.07 & 1.19 & 0.25 & 0.51 & 0.79 & 0.92 & 0.34 & 0.67 & 1.22 & 1.47 & 0.60 & 0.98 & 1.36 & 1.50 & 0.33 & 0.66 & 0.94 & 0.99 & 0.40 & 0.78 & 1.07 & 1.21\\
  Residual sup. \cite{Martinez_2017_CVPR}& 0.39 & 0.81 & 1.40 & 1.62 & 0.24 & 0.51 & 0.90 & 1.05 & 0.28 & 0.53 & 1.02 & 1.14 & 0.56 & 0.91 & 1.26 & 1.40 & 0.31 & 0.58 & 0.87 & 0.91 & 0.36 & 0.67 & 1.02 & 1.15 \\
  convSeq2Seq \cite{LiZLL18} & 0.41 & 0.78 & 1.16 & 1.31 & 0.23 & 0.49 & 0.88 & 1.06 & 0.30 & 0.62 & 1.09 & 1.30 & 0.59 & 1.00 & 1.32 & 1.44 & 0.27 & 0.52 & 0.71 & 0.74  & 0.38 & 0.68 & 1.01 & 1.13\\
  AGED w/o adv \cite{gui2018adversarial}& 0.38 & 0.77 & 1.18 & 1.41 & 0.24 & 0.52 & 0.92 & 1.01 & 0.31 & 0.64 & 1.08 & 1.12 & 0.51 & 0.87 & 1.21 & 1.33 & 0.29 & 0.51 & 0.72 & 0.75 & 0.32 & 0.62 & 0.96 & 1.07\\
  AGED w/adv \cite{gui2018adversarial}& 0.33 & 0.62 & 0.98 & 1.1 & 0.23 & 0.48 & 0.81 & 0.95 & 0.24 & \textbf{0.50} & 1.02 & \textbf{1.13} & 0.50 & 0.81 & 1.15 & \textbf{1.27} & 0.23 & 0.41 & 0.56 & 0.62 & 0.31 & 0.54 & 0.85 & 0.97 \\\hline
  Ours & \textbf{0.30} & \textbf{0.61} & \textbf{0.90} & \textbf{1.00} & \textbf{0.14} & \textbf{0.34} & \textbf{0.58} & \textbf{0.70} & \textbf{0.23} & \textbf{0.50} & \textbf{0.91} & 1.14 & \textbf{0.46} & \textbf{0.79} & \textbf{1.12} & 1.29 & \textbf{0.15} & \textbf{0.34} & \textbf{0.52} & \textbf{0.57} & \textbf{0.27} & \textbf{0.51} & \textbf{0.83} & \textbf{0.95}\\\hline
\end{tabular}
}
\caption{Short-term prediction of joint angles on H3.6M for all actions. Our method outperforms the state of the art for most time horizons.
}
\vspace{-0.35cm}
\label{table-short-motion}
\end{table*}
\begin{table*}[!ht]
\centering
\resizebox{0.95\textwidth}{!}{%
\begin{tabular}{ccccc|cccc|cccc|cccc}
  & \multicolumn{4}{c}{Walking} & \multicolumn{4}{c}{Eating} & \multicolumn{4}{c}{Smoking} & \multicolumn{4}{c}{Discussion}\\
  milliseconds& 80 & 160 & 320 & 400 & 80 & 160 & 320 & 400 & 80 & 160 & 320 & 400 & 80 & 160 & 320 & 400\\ \hline
Residual sup. \cite{Martinez_2017_CVPR}& 21.7 & 38.1 & 58.9 & 68.8 & 15.1 & 28.6 & 54.8 & 67.4 & 20.8 & 39.0 & 66.1 & 76.1 & 26.2 & 51.2 & 85.8 & 94.6 \\
Residual sup. 3D\cite{Martinez_2017_CVPR} & 23.8 & 40.4 & 62.9 & 70.9 & 17.6 & 34.7 & 71.9 & 87.7 & 19.7 & 36.6 & 61.8 & 73.9 & 31.7 & 61.3 & 96.0 & 103.5 \\
convSeq2Seq \cite{LiZLL18}& 21.8 & 37.5 & 55.9 & 63.0 & 13.3 & 24.5 & 48.6 & 60.0 & 15.4 & 25.5 & 39.3 & 44.5 & 23.6 & 43.6 & 68.4 & 74.9\\
convSeq2Seq 3D \cite{LiZLL18} & 17.1 & 31.2 & 53.8 & 61.5 & 13.7 & 25.9 & 52.5 & 63.3 & 11.1 & 21.0 & 33.4 & 38.3 & 18.9 & 39.3 & 67.7 & 75.7 \\\hline
Ours& 11.1 & 19.0 & 32.0 & 39.1 & 9.2 & 19.5 & 40.3 & 48.9 & 9.2 & 16.6 & 26.1 & 29.0 & 11.3 & 23.7 & 41.9 & 46.6 \\
Ours 3D & \textbf{8.9} & \textbf{15.7} & \textbf{29.2} & \textbf{33.4} & \textbf{8.8} & \textbf{18.9} & \textbf{39.4} & \textbf{47.2} & \textbf{7.8} & \textbf{14.9} & \textbf{25.3} & \textbf{28.7} & \textbf{9.8} & \textbf{22.1} & \textbf{39.6} & \textbf{44.1}\\\hline
\end{tabular}
}
\resizebox{\textwidth}{!}{%
\begin{tabular}{ccccc|cccc|cccc|cccc|cccc|cccc}
  & \multicolumn{4}{c}{Directions} & \multicolumn{4}{c}{Greeting} & \multicolumn{4}{c}{Phoning} & \multicolumn{4}{c}{Posing}&\multicolumn{4}{c}{Purchases}&\multicolumn{4}{c}{Sitting} \\
  milliseconds& 80 & 160 & 320 & 400 & 80 & 160 & 320 & 400 & 80 & 160 & 320 & 400 & 80 & 160 & 320 & 400& 80 & 160 & 320 & 400& 80 & 160 & 320 & 400 \\ \hline
Residual sup.\cite{Martinez_2017_CVPR} & 27.9 & 44.8 & 63.5 & 78.2 & 29.3 & 56.0 & 110.2 & 125.6 & 28.7 & 50.9 & 88.0 & 99.7 & 30.5 & 59.4 & 118.7 & 144.7 & 33.3 & 58.2 & 85.4 & 93.7 & 32.6 & 65.2 & 113.7 & 126.2\\
  Residual sup. 3D \cite{Martinez_2017_CVPR} & 36.5 & 56.4 & 81.5 & 97.3 & 37.9 & 74.1 & 139.0 & 158.8 & 25.6 & 44.4 & 74.0 & 84.2 & 27.9 & 54.7 & 131.3 & 160.8 & 40.8 & 71.8 & 104.2 & 109.8 & 34.5 & 69.9 & 126.3 & 141.6\\
  convSeq2Seq\cite{LiZLL18} & 26.7 & 43.3 & 59.0 & 72.4 & 30.4 & 58.6 & 110.0 & 122.8 & 22.4 & 38.4 & 65.0 & 75.4 & 22.4 & 42.1 & 87.3 & 106.1 & 28.4 & 53.8 & 82.1 & 93.1 & 24.7 & 50.0 & 88.6 & 100.4\\
  convSeq2Seq 3D \cite{LiZLL18} & 22.0 & 37.2 & 59.6 & 73.4 & 24.5 & 46.2 & 90.0 & 103.1 & 17.2 & 29.7 & 53.4 & 61.3 & 16.1 & 35.6 & 86.2 & 105.6 & 29.4 & 54.9 & 82.2 & 93.0 & 19.8 & 42.4 & 77.0 & 88.4\\\hline
 Ours & \textbf{11.2} & \textbf{23.2} & 52.7 & 64.1 & \textbf{14.2} & \textbf{27.7} & \textbf{67.1} & \textbf{82.9} & 13.5 & 22.5 & 45.2 & 52.4 & 11.1 & 27.1 & 69.4 & 86.2 & 20.4 & 42.8 & 69.1 & 78.3 & 11.7 & 27.0 & 55.9 & 66.9\\
 Ours 3D & 12.6 & 24.4 & \textbf{48.2} & \textbf{58.4} & 14.5 & 30.5 & 74.2 & 89.0 & \textbf{11.5} & \textbf{20.2} & \textbf{37.9} & \textbf{43.2} & \textbf{9.4} & \textbf{23.9} & \textbf{66.2} & \textbf{82.9} & \textbf{19.6} & \textbf{38.5} & \textbf{64.4} & \textbf{72.2} & \textbf{10.7} & \textbf{24.6} & \textbf{50.6} & \textbf{62.0} \\ \hline
  & \multicolumn{4}{c}{Sitting Down} & \multicolumn{4}{c}{Taking Photo} & \multicolumn{4}{c}{Waiting} & \multicolumn{4}{c}{Walking Dog}&\multicolumn{4}{c}{Walking Together}&\multicolumn{4}{c}{Average} \\
  milliseconds& 80 & 160 & 320 & 400 & 80 & 160 & 320 & 400 & 80 & 160 & 320 & 400 & 80 & 160 & 320 & 400& 80 & 160 & 320 & 400& 80 & 160 & 320 & 400 \\ \hline
  Residual sup. \cite{Martinez_2017_CVPR}   & 33.0 & 64.1 & 121.7 & 146 & 21.2 & 40.3 & 72.2 & 86.2 & 24.9 & 50.0 & 96.5 & 114.0 & 53.8 & 90.9 & 134.6 & 156.9 & 19.7 & 38.2 & 62.9 & 72.3 & 27.9 & 51.6 & 88.9 & 103.4\\
  Residual sup. 3D \cite{Martinez_2017_CVPR} & 28.6 & 55.3 & 101.6 & 118.9 & 23.6 & 47.4 & 94.0 & 112.7 & 29.5 & 60.5 & 119.9 & 140.6 & 60.5 & 101.9 & 160.8 & 188.3 & 23.5 & 45.0 & 71.3 & 82.8 & 30.8 & 57.0 & 99.8 & 115.5\\
  convSeq2Seq \cite{LiZLL18} & 23.9 & 39.9 & 74.6 & 89.8 & 18.4 & 32.1 & 60.3 & 72.5 & 24.9 & 50.2 & 101.6 & 120.0 & 56.4 & 94.9 & 136.1 & 156.3 & 21.1 & 38.5 & 61.0 & 70.4 & 24.9 & 44.9 & 75.9 & 88.1\\
  convSeq2Seq 3D \cite{LiZLL18} & 17.1 & 34.9 & 66.3 & 77.7 & 14.0 & 27.2 & 53.8 & 66.2 & 17.9 & 36.5 & 74.9 & 90.7 & 40.6 & 74.7 & 116.6 & 138.7 & 15.0 & 29.9 & 54.3 & 65.8 & 19.6 & 37.8 & 68.1 & 80.2\\\hline
 Ours & 11.5 & \textbf{25.4} & \textbf{53.9} & \textbf{65.6} & 8.3 & 15.8 & 38.5 & \textbf{49.1} & 12.1 & 27.5 & 67.3 & 85.6 & 35.8 & 63.6 & 106.7 & 126.8 & 11.7 & 23.5 & 46.0 & 53.5 & 13.5 & 27.0 & 54.2 & 65.0\\
 Ours 3D & \textbf{11.4} & 27.6 & 56.4 & 67.6 & \textbf{6.8} & \textbf{15.2} & \textbf{38.2} & 49.6 & \textbf{9.5} & \textbf{22.0} & \textbf{57.5} & \textbf{73.9} & \textbf{32.2} & \textbf{58.0} & \textbf{102.2} & \textbf{122.7} & \textbf{8.9} & \textbf{18.4} & \textbf{35.3} & \textbf{44.3} & \textbf{12.1} & \textbf{25.0} & \textbf{51.0} & \textbf{61.3}\\\hline
\end{tabular}
}
\caption{Short-term prediction of 3D joint positions on H3.6M. A {\it 3D} in the method's name indicates that it was directly trained on 3D joint positions. Otherwise, the results were obtained by converting the angle predictions to 3D positions. Note that we outperform the baselines by a large margin, particularly when training directly on 3D.
}
\vspace{-0.55cm}
\label{table-short-3d}
\end{table*}

%% file: 4_experiments.tex
\section{Experiments}
We evaluate our model on several benchmark motion capture (mocap) datasets, including Human3.6M (H3.6M)~\cite{h36m_pami}, the CMU mocap dataset\footnote{Available at \url{http://mocap.cs.cmu.edu/}}, and the 3DPW dataset~\cite{vonMarcard2018}. Below, we first introduce these datasets, the evaluation metrics we use and the baselines we compare our method with. We then present our results using both joint angles and 3D coordinates.

\begin{figure*}[ht]
    \centering
    \begin{tabular}{cc}
      \includegraphics[width=0.46\linewidth]{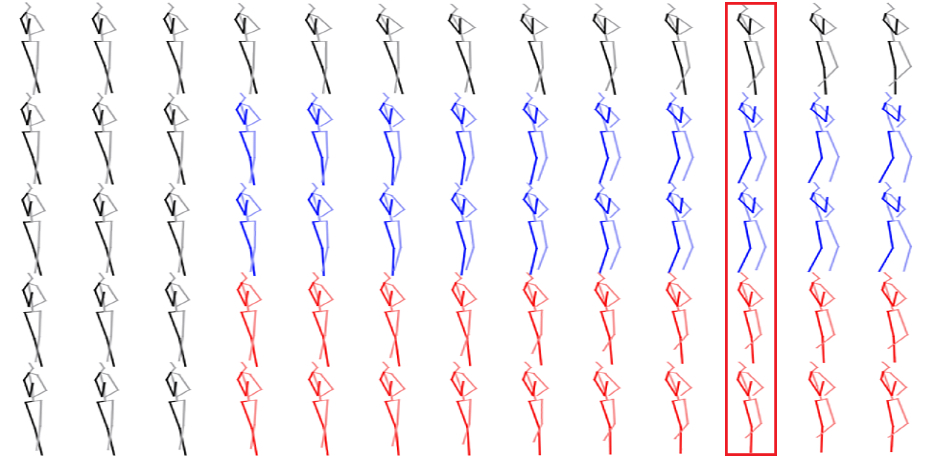} & 
      \includegraphics[width=0.46\linewidth]{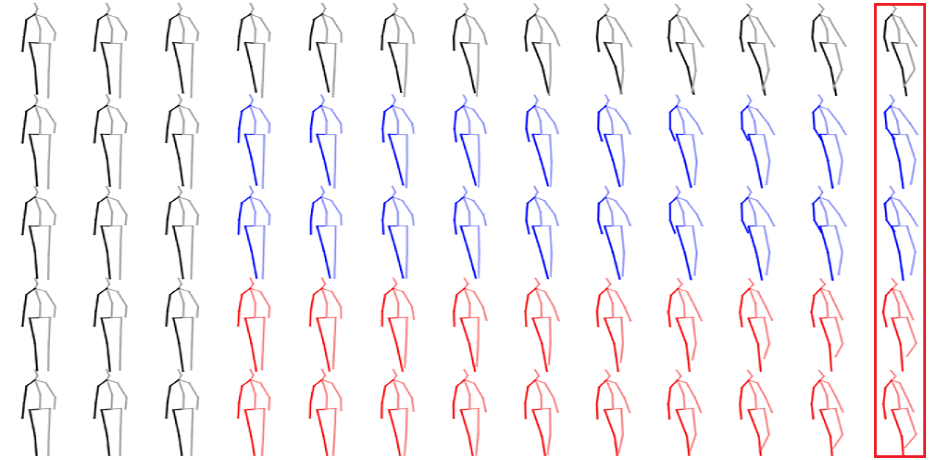} \\
      (a)Smoking & (b)Walking\\
      \multicolumn{2}{c}{\includegraphics[width=0.96\linewidth]{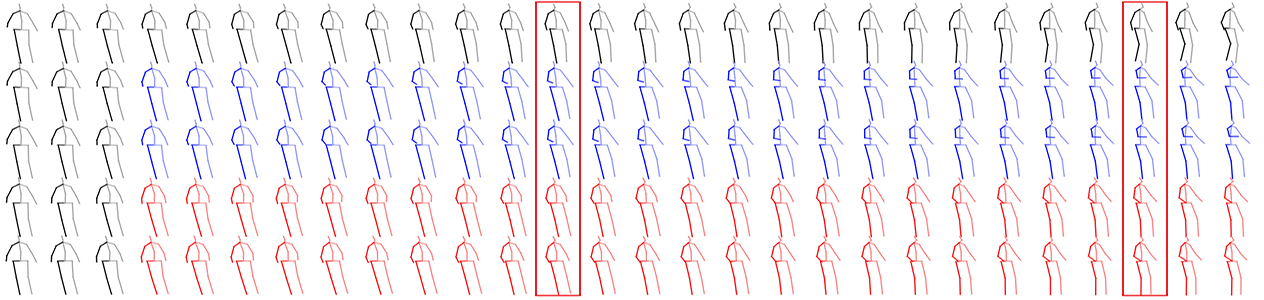}}\\
      \multicolumn{2}{c}{(c)Walking Dog}
    \end{tabular}
    \caption{Qualitative comparison of short-term (``Smoking" and ``Walking") and long-term (``Walking Dog") predictions on H3.6M. From top to bottom, we show the ground truth, and the results of Residual sup.~\cite{Martinez_2017_CVPR}, convSeq2Seq~\cite{LiZLL18}, our approach based on angles, and our approach based on 3D positions. The results evidence that our approach generates high-quality predictions in both cases.}
    \vspace{-0.5cm}
    \label{fig:short-pred-quali-h36m}
\end{figure*}

\subsection{Datasets}
\noindent\textbf{Human3.6M.} To the best of our knowledge, Human3.6M (H3.6M)\cite{h36m_pami} is the largest dataset for human motion analysis. It depicts seven actors performing 15 actions, such as walking, eating, discussion, sitting, and phoning. The actors are represented by a skeleton of 32 joints. Following the data processing of~\cite{gui2018adversarial,Martinez_2017_CVPR}, we remove the global rotations and translations as well as constant angles. 
\comment{because this information is independent of the activity itself. We also remove the angles with constant standard deviation. This leaves us with pose vectors of dimension 48.~It has 66 joint 3D coordinates in total by ignoring a few with fixed 3D coordinates such as pelvicum and left/right hip across the motion frames.}The sequences are down-sampled to 25 frames per second and we test on the same sequences of subject 5 (S5) as previous work \cite{gui2018adversarial,LiZLL18,Martinez_2017_CVPR}. \comment{In contrast to previous work \cite{gui2018adversarial,LiZLL18,Martinez_2017_CVPR}, we keep subject 11 (S11) as validation data to choose the best model and use the remaining 5 subjects (S1,S6,S7,S8,S9) as training set.}

\noindent\textbf{CMU-Mocap.} Following~\cite{LiZLL18}, we also report results
on the CMU mocap dataset (CMU-Mocap). For a fair comparison, we adopt the same data representation and training/test splits as in~\cite{LiZLL18}, provided in their released code and data. Based on~\cite{LiZLL18}, eight actions are selected for evaluation after pre-processing the entire dataset by removing sequences depicting multiple people, sequences with less training data and actions with repetitions. We apply the same pre-processing as on H3.6M.
\comment{In this dataset, the skeleton has 38 joints and 114 joint angles in total. After removing global translation, rotation and those with constant standard deviation, it leaves us 64 joint angles.~Following the processing procedure as H3.6M for 3D joint coordinate, we use 75 joint coordinates. We do not split a validation set from the given dataset due to limited training data.}

\begin{table}[!ht]
\centering
\resizebox{\linewidth}{!}{%
\begin{tabular}{ccc|cc|cc|cc|cc}
  & \multicolumn{2}{c}{Walking} & \multicolumn{2}{c}{Eating} & \multicolumn{2}{c}{Smoking} & \multicolumn{2}{c}{Discussion}&\multicolumn{2}{c}{Average} \\
  milliseconds& 560 & 1000 & 560 & 1000 & 560 & 1000 & 560 & 1000& 560 & 1000 \\ \hline
  zero-velocity \cite{Martinez_2017_CVPR} & 1.35 & 1.32 & 1.04 & 1.38 & 1.02 & 1.69 & 1.41 & 1.96 & 1.21 & 1.59\\
Residual sup.  \cite{Martinez_2017_CVPR} & 0.93 & 1.03 & 0.95 & 1.08 & 1.25 & 1.50 & 1.43 & 1.69 & 1.14 & 1.33\\
convSeq2Seq \cite{LiZLL18} & N/A & 0.92 & N/A & 1.24 & N/A & 1.62 & N/A & 1.86 & N/A & 1.41\\
AGED w/o adv \cite{gui2018adversarial} & 0.89 & 1.02 & 0.92 & 1.01 & 1.15 & 1.43 & 1.33 & 1.5 & 1.07 & 1.24\\
AGED w/adv \cite{gui2018adversarial} & 0.78 & 0.91 & 0.86 & \textbf{0.93} & 1.06 & \textbf{1.21} & \textbf{1.25} & \textbf{1.30} & 0.99 & \textbf{1.09}\\\hline
Ours & \textbf{0.65} & \textbf{0.67} & \textbf{0.76} & 1.12 & \textbf{0.87} & 1.57 & 1.33 & 1.70 & \textbf{0.90} & 1.27\\
  \hline\hline
  Residual sup. \cite{Martinez_2017_CVPR} & 79.4 & 91.6 & 82.6 & 110.8 & 89.5 & 122.6 & 121.9 & 154.3 & 93.3 & 119.8\\
Residual sup. 3D \cite{Martinez_2017_CVPR} & 73.8 & 86.7 & 101.3 & 119.7 & 85.0 & 118.5 & 120.7 & 147.6 & 95.2 & 118.1\\
convSeq2Seq \cite{LiZLL18} & 69.2 & 81.5 & 71.8 & 91.4 & 50.3 & 85.2 & 101.0 & 143.0 & 73.1 & 100.3\\
convSeq2Seq 3D\cite{LiZLL18} & 59.2 & 71.3 & 66.5 & 85.4 & 42.0 & 67.9 & 84.1 & 116.9 & 62.9 & 85.4\\\hline
  Ours & 55.0 & 60.8 & 68.1 & 79.5 & 42.2 & 70.6 & 93.8 & 119.7 & 64.8 & 82.6\\
Ours 3D & \textbf{42.3} & \textbf{51.3} & \textbf{56.5} & \textbf{68.6} & \textbf{32.3} & \textbf{60.5} & \textbf{70.5} & \textbf{103.5} &   \textbf{50.4} & \textbf{71.0}\\\hline
\end{tabular}
}
\caption{Long-term prediction of joint angles (top) and 3D joint positions (bottom) on H3.6M.}
\vspace{-0.6cm}
\label{table-long-motion}
\end{table}
\noindent\textbf{3DPW.} The 3D Pose in the Wild dataset (3DPW)~\cite{vonMarcard2018} is a recently published dataset which has more than 51k frames with 3D annotations for challenging indoor and outdoor activities. We use the official training, test and validation sets. The frame rate of the 3D annotation is 30Hz. 
\comment{The human skeleton has 24 joints with 72 joint angles. It has 69 joint coordinate parameters by ignoring the joints with fixed 3D coordinates after removing global translation.}
\subsection{Evaluation Metrics and Baselines}
\begin{table*}[ht]
\centering
\resizebox{\textwidth}{!}{%
\begin{tabular}{cccccc|ccccc|ccccc|ccccc|ccccc}
  & \multicolumn{5}{c}{Basketball} & \multicolumn{5}{c}{Basketball Signal} & \multicolumn{5}{c}{Directing Traffic} & \multicolumn{5}{c}{Jumping} & \multicolumn{5}{c}{Running} \\
  milliseconds& 80 & 160 & 320 & 400 & 1000 & 80 & 160 & 320 & 400 & 1000 & 80 & 160 & 320 & 400 & 1000 & 80 & 160 & 320 & 400 & 1000 & 80 & 160 & 320 & 400 & 1000 \\ \hline
 Residual sup. \cite{Martinez_2017_CVPR} & 0.50 & 0.80 & 1.27 & 1.45 & 1.78 & 0.41 & 0.76 & 1.32 & 1.54 & 2.15 & 0.33 & 0.59 & 0.93 & 1.10 & 2.05 & 0.56 & 0.88 & 1.77 & 2.02 & 2.4 & 0.33 & 0.50 & 0.66 & 0.75 & 1.00\\
 convSeq2Seq \cite{LiZLL18} & 0.37 & 0.62 & 1.07 & 1.18 & 1.95 & 0.32 & 0.59 & 1.04 & 1.24 & 1.96 & 0.25 & 0.56 & 0.89 & 1.00 & 2.04 & 0.39 & 0.6 & 1.36 & 1.56 & 2.01 & \textbf{0.28} & \textbf{0.41} & \textbf{0.52} & \textbf{0.57} & \textbf{0.67}\\\hline
 Ours & \textbf{0.33} & \textbf{0.52} & \textbf{0.89} & \textbf{1.06} & \textbf{1.71} & \textbf{0.11} & \textbf{0.20} & \textbf{0.41} & \textbf{0.53} & \textbf{1.00} & \textbf{0.15} & \textbf{0.32} & \textbf{0.52} & \textbf{0.60} & \textbf{2.00} & \textbf{0.31} & \textbf{0.49} & \textbf{1.23} & \textbf{1.39} & \textbf{1.80} & 0.33 & 0.55 & 0.73 & 0.74 & 0.95\\\hline
  & \multicolumn{5}{c}{Soccer} & \multicolumn{5}{c}{Walking} & \multicolumn{5}{c}{Washwindow} & \multicolumn{5}{c}{Average} & \\
  milliseconds& 80 & 160 & 320 & 400 & 1000 & 80 & 160 & 320 & 400 & 1000 & 80 & 160 & 320 & 400 & 1000 & 80 & 160 & 320 & 400 & 1000 \\ \cline{1-21}
 Residual sup. \cite{Martinez_2017_CVPR} & 0.29 & 0.51 & 0.88 & 0.99 & 1.72 & 0.35 & 0.47 & 0.60 & 0.65 & 0.88 & 0.30 & 0.46 & 0.72 & 0.91 & 1.36 & 0.38 & 0.62 & 1.02 & 1.18 & 1.67\\
 convSeq2Seq \cite{LiZLL18} & 
0.26 & 0.44 & 0.75 & 0.87 & 1.56 & 0.35 & \textbf{0.44} & \textbf{0.45} & \textbf{0.50} & 0.78 & 0.30 & 0.47 & 0.80 & 1.01 & 1.39 & 0.32 & 0.52 & 0.86 & 0.99 & 1.55\\\cline{1-21}
 Ours & 
\textbf{0.18} & \textbf{0.29} & \textbf{0.61} & \textbf{0.71} & \textbf{1.40} & \textbf{0.33} & 0.45 & 0.49 & 0.53 & \textbf{0.61} & \textbf{0.22} & \textbf{0.33} & \textbf{0.57} & \textbf{0.75} & \textbf{1.20} & \textbf{0.25} & \textbf{0.39} & \textbf{0.68} & \textbf{0.79} & \textbf{1.33} \\\cline{1-21}
\hline\hline
& \multicolumn{5}{c}{Basketball} & \multicolumn{5}{c}{Basketball Signal} & \multicolumn{5}{c}{Directing Traffic} & \multicolumn{5}{c}{Jumping} & \multicolumn{5}{c}{Running} \\
  milliseconds& 80 & 160 & 320 & 400 & 1000 & 80 & 160 & 320 & 400 & 1000 & 80 & 160 & 320 & 400 & 1000 & 80 & 160 & 320 & 400 & 1000 & 80 & 160 & 320 & 400 & 1000 \\ \hline
 Residual sup. 3D\cite{Martinez_2017_CVPR} &18.4 & 33.8 & 59.5 & 70.5 & 106.7 & 12.7 & 23.8 & 40.3 & 46.7 & 77.5 & 15.2 & 29.6 & 55.1 & 66.1 & 127.1 & 36.0 & 68.7 & 125.0 & 145.5 & 195.5 & 15.6 & 19.4 & 31.2 & 36.2 & 43.3\\
 convSeq2Seq 3D\cite{LiZLL18} &16.7 & 30.5 & 53.8 & 64.3 & \textbf{91.5} & 8.4 & 16.2 & 30.8 & 37.8 & 76.5 & 10.6 & 20.3 & 38.7 & 48.4 & \textbf{115.5} & 22.4 & 44.0 & 87.5 & 106.3 & \textbf{162.6} & \textbf{14.3} & \textbf{16.3} & \textbf{18.0} & \textbf{20.2} & \textbf{27.5}\\
 \hline
 Ours 3D& \textbf{14.0} & \textbf{25.4} & \textbf{49.6} & \textbf{61.4} & 106.1 & \textbf{3.5} & \textbf{6.1} & \textbf{11.7} & \textbf{15.2} & \textbf{53.9} & \textbf{7.4} & \textbf{15.1} & \textbf{31.7} & \textbf{42.2} & 152.4 & \textbf{16.9} & \textbf{34.4} & \textbf{76.3} & \textbf{96.8} & 164.6 & 25.5 & 36.7 & 39.3 & 39.9 & 58.2\\\hline
  & \multicolumn{5}{c}{Soccer} & \multicolumn{5}{c}{Walking} & \multicolumn{5}{c}{Washwindow} & \multicolumn{5}{c}{Average} & \\
  milliseconds& 80 & 160 & 320 & 400 & 1000 & 80 & 160 & 320 & 400 & 1000 & 80 & 160 & 320 & 400 & 1000 & 80 & 160 & 320 & 400 & 1000 & & & & & \\ \cline{1-21}
  Residual sup. 3D\cite{Martinez_2017_CVPR} & 20.3 & 39.5 & 71.3 & 84 & 129.6 & 8.2 & 13.7 & 21.9 & 24.5 & 32.2 & 8.4 & 15.8 & 29.3 & 35.4 & 61.1 & 16.8 & 30.5 & 54.2 & 63.6 & 96.6\\
convSeq2Seq 3D\cite{LiZLL18} & 12.1 & 21.8 & \textbf{41.9} & \textbf{52.9} & \textbf{94.6} & \textbf{7.6} & 12.5 & 23.0 & 27.5 & 49.8 & 8.2 & 15.9 & 32.1 & \textbf{39.9} & \textbf{58.9} & 12.5 & 22.2 & 40.7 & 49.7 & \textbf{84.6}\\\cline{1-21}
Ours 3D& \textbf{11.3} & \textbf{21.5} & 44.2 & 55.8 & 117.5 & 7.7 & \textbf{11.8} & \textbf{19.4} & \textbf{23.1} & \textbf{40.2} & \textbf{5.9} & \textbf{11.9} & \textbf{30.3} & 40.0 & 79.3 & \textbf{11.5} & \textbf{20.4} & \textbf{37.8} & \textbf{46.8} & 96.5\\\cline{1-21}
\end{tabular}
}
\caption{Short and long-term prediction of joint angles (top) and 3D joint positions (bottom) on CMU-Mocap.}
\vspace{-0.4cm}
\label{table-short-motion-cmu}
\end{table*}
\begin{table}[!ht]
\centering
\resizebox{0.4\textwidth}{!}{%
\begin{tabular}{c|ccccc}
  milliseconds& 200 & 400 & 600 & 800 & 1000\\ \hline
Residual sup.~\cite{Martinez_2017_CVPR} & 1.85 & 2.37 & 2.46 & 2.51 & 2.53 \\
convSeq2Seq~\cite{LiZLL18} & 1.24 & 1.85 &  2.13 & 2.23 & 2.26 \\\hline
Ours & \textbf{0.64} & \textbf{0.95} & \textbf{1.12} & \textbf{1.22} & \textbf{1.27}\\\hline\hline
Residual sup. 3D~\cite{Martinez_2017_CVPR} & 113.9 & 173.1 & 191.9 & 201.1 & 210.7 \\
convSeq2Seq 3D~\cite{LiZLL18} & 71.6 & 124.9 & 155.4 & 174.7 & 187.5 \\\hline
Ours 3D & \textbf{35.6} & \textbf{67.8} & \textbf{90.6} & \textbf{106.9} & \textbf{117.8}\\\hline
\end{tabular}
}
\caption{Short-term and long-term prediction of joint angle (top) and 3D joint positions (bottom) on 3DPW.}
\vspace{-0.3cm}
\label{table-short-long-3dpw}
\end{table}
\noindent\textbf{Metrics.} We follow the standard evaluation protocol used in~\cite{Martinez_2017_CVPR,LiZLL18,gui2018adversarial}, and report the Euclidean distance between the predicted and ground-truth joint angles in Euler angle representation. We further report results in terms of 3D error. To this end, we make use of the Mean Per Joint Position Error (MPJPE)~\cite{h36m_pami} in millimeter, commonly used for image-based 3D human pose estimation.

As will be shown later, 3D errors can be measured either by directly train a model on the 3D coordinates (via the DCT in our case), or by converting the predicted angles to 3D. 

\noindent\textbf{Baselines.}
We compare our approach with two recent RNN-based methods, namely, Residual sup.~\cite{Martinez_2017_CVPR} and AGED (w or w/o adv)~\cite{gui2018adversarial}, and with one feedforward model, convSeq2Seq~\cite{LiZLL18}. When reporting angular errors, we directly make use of the results provided in the respective papers of these baselines. Because these works do not report 3D error, in this case, we rely on the code provided by the authors of~\cite{Martinez_2017_CVPR,LiZLL18}, which we adapted so as to take 3D coordinates as input and output. Note that the code of~\cite{gui2018adversarial} is not available, and we were unable to reproduce their method so as to obtain reliable results with their adversarial training strategy\footnote{Note that the geodesic loss of~\cite{gui2018adversarial} does not apply to 3D space.}. Therefore, we only report the results of this method in angle space.

\noindent\textbf{Implementation details.}
We implemented our network using Pytorch~\cite{paszke2017automatic}, and we used ADAM~\cite{kingma2014adam} to train our model.
The learning rate was set to 0.0005 with a 0.96 decay every two epochs. The batch size was set to 16 and the gradients were clipped to a maximum $\ell$2-norm of 1. It takes 30ms for one forward pass and back-propagation on an NVIDIA Titan V GPU. Our models are trained for 50 epochs. More details about the experiments are included in the supplementary material.
\comment{and choose the last model for CMU-Mocap dataset. We chose the best one based on the validation set for H3.6M dataset and 3DPW dataset.}

\subsection{Results}
To be consistent with the literature, we report our results for short-term ($<500ms$)  and long-term ($>500ms$) predictions. For all datasets, we are given $10$ frames (400 milliseconds) to predict the future $10$ frames (400 milliseconds) for short-term prediction and to predict the future $25$ frames (1 second) for long-term prediction.

\noindent\textbf{Human 3.6M.} In Table~\ref{table-short-motion}, we compare our results to those of the baselines for short-term prediction in angle space on H3.6M. Table~\ref{table-short-motion} reports the errors for the activities ``Walking", ``Eating", ``Smoking" and ``Discussion", which have been the focus of the comparisons in the literature. It also provides the results for the other 11 activities and the average over the 15 activities. Note that we outperform all the baselines on average. 
We provide qualitative comparisons in Fig.~\ref{fig:short-pred-quali-h36m}. They further evidence that our predictions are closer to the ground truth than that of the baselines for all 3 actions. More visualizations are included in the supplementary material.

\begin{figure}[!ht]
    \hspace{-0.4cm}
    \begin{tabular}{c}
          \includegraphics[width=0.90\linewidth]{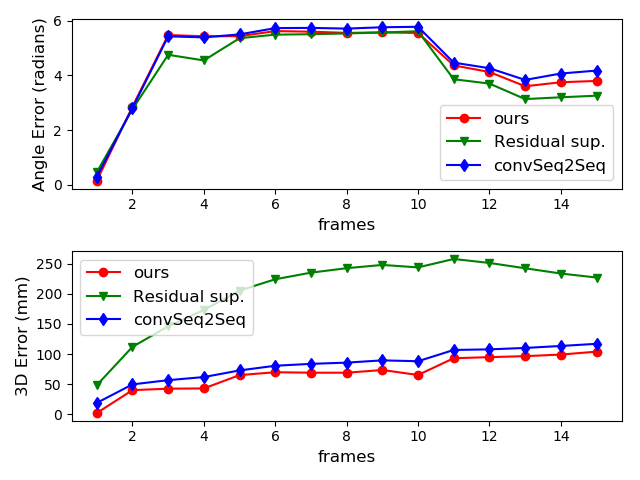} \vspace{-0.2cm}\\
          (a)\\
         \includegraphics[width=0.95\linewidth]{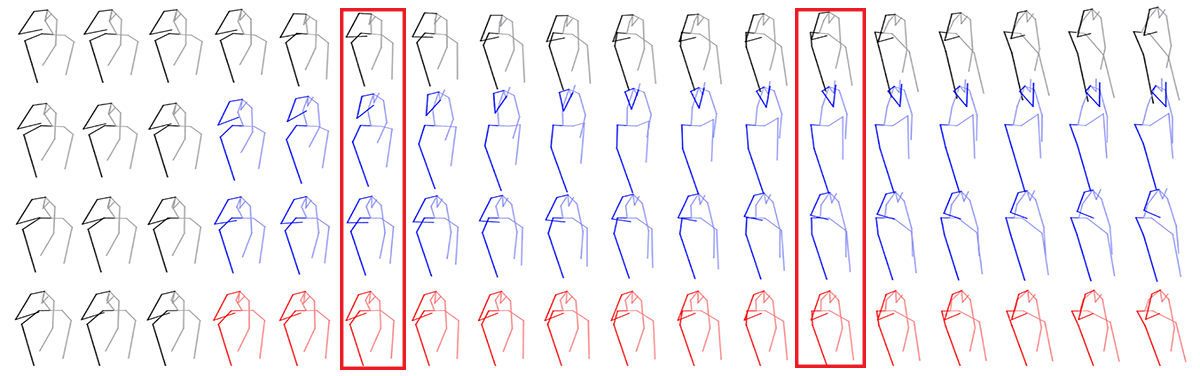} \\
          (b)
    \end{tabular}
    \caption{Drawbacks of the angle-based representation. (a) Joint angle error (top) and 3D position error (bottom) for each predicted frame on the Phoning H3.6M action. While all methods have a similar error in angle space, Residual sup.~\cite{Martinez_2017_CVPR} yields a  much higher one in 3D.
    This is also reflected by the qualitative comparison in (b). In the predictions of~\cite{Martinez_2017_CVPR} (2nd row), the 3D location of the right hand and left leg are too high and far away from the ground truth, leading to unrealistic poses. By contrast, the predictions of~\cite{LiZLL18} (3rd row) and our method (last row) are closer to the ground truth.}
    \vspace{-0.5cm}
    \label{fig:angle-3d-ambi}
\end{figure}

To analyze the failure cases of our approach, such as for ``Phoning", we converted the predicted angles to 3D coordinates so as to visualize the poses. We were then surprised to realize that a high error in angle space did \emph{not} necessarily translate to a high error in 3D space. This is due to the fact that the angle representation is ambiguous, and thus two very different sets of angles can yield the same pose. To evidence this, in Fig.~\ref{fig:angle-3d-ambi}, we plot the angle error for three methods, including ours, on the same sequence, as well as the corresponding 3D errors obtained by simply converting the angles to 3D coordinates. Note that, while all three methods have comparable errors in angle space, two of them, including ours, have a \emph{much} lower error than the third one in 3D space. This makes us argue that angles are not a good representation to evaluate motion prediction.

Motivated by this observation, in Table~\ref{table-short-3d}, we report the 3D errors for short-term prediction on H3.6M. As mentioned before, there are two ways to achieve this: Converting the predicted angles to 3D or directly training the models on 3D coordinates. We report the results of both strategies. Note that, having access to neither the code nor the angle predictions of~\cite{gui2018adversarial}, we are unable to provide the 3D results for this method. When considering the remaining baselines, our approach consistently outperforms them, yielding the best results when directly using the 3D information (via the DCT) during training.
\begin{table*}[ht]
\centering
\resizebox{\linewidth}{!}{%
\begin{tabular}{ccc|cccc|cccc|cccc|cccc|cccc}
  \multicolumn{3}{c}{}& \multicolumn{4}{c}{Walking} & \multicolumn{4}{c}{Eating} & \multicolumn{4}{c}{Smoking} & \multicolumn{4}{c}{Discussion}&\multicolumn{4}{c}{Average} \\
  dct&padding&resi& 80 & 160 & 320 & 400 & 80 & 160 & 320 & 400 & 80 & 160 & 320 & 400 & 80 & 160 & 320 & 400& 80 & 160 & 320 & 400 \\ \hline
  & \checkmark &\checkmark & 0.20 & 0.33 & 0.52 & 0.59 & 0.17 & 0.30 & \textbf{0.50} & \textbf{0.62} & \textbf{0.22} & \textbf{0.41} & \textbf{0.83} & \textbf{0.78} & 0.24 & 0.60 & 0.91 & 0.97 & 0.21 & 0.41 & 0.69 & 0.74 \\
 \checkmark &  & \checkmark & 0.34 & 0.46 & 0.65 & 0.71 & 0.33 & 0.44 & 0.63 & 0.76 & 0.47 & 0.60 & 0.94 & 0.95 & 0.40 & 0.70 & 0.95 & 1.00 & 0.39 & 0.55 & 0.79 & 0.86 \\
 \checkmark & \checkmark &  & 0.25 & 0.41 & 0.62 & 0.69 & 0.26 & 0.39 & 0.60 & 0.73 & 0.31 & 0.49 & 0.89 & 0.89 & 0.34 & 0.72 & 0.97 & 1.02 & 0.29 & 0.50 & 0.77 & 0.83\\
 \checkmark & \checkmark & \checkmark & \textbf{0.18} & \textbf{0.31} & \textbf{0.49} & \textbf{0.56} & \textbf{0.16} & \textbf{0.29} & \textbf{0.50} & \textbf{0.62} & \textbf{0.22} & \textbf{0.41} & 0.86 & 0.80 & \textbf{0.20} & \textbf{0.51} & \textbf{0.77} & \textbf{0.85} & \textbf{0.19} & \textbf{0.38} & \textbf{0.66} & \textbf{0.71} \\
 \hline
 \hline
  & \checkmark &\checkmark & 11.4 & 19.5 & 32.9 & 38.3 & 10.6 & 21.4 & 41.1 & 48.0 & 9.4 & 16.7 & 27.2 & 32.2 & 14.1 & 29.6 & 49.9 & 54.1 & 11.4 & 21.8 & 37.8 & 43.1\\
 \checkmark &  & \checkmark & 19.1 & 24.7 & 37.3 & 41.5 & 24.7 & 30.4 & 48.6 & 55.8 & 40.5 & 41.0 & 48.9 & 53.0 & 22.6& 29.9 & 46.7 & 51.3 & 26.7 & 31.5 & 45.4 & 50.4\\
 \checkmark & \checkmark &  & 18.3 & 25.9 & 39.7 & 43.7 & 20.1& 29.4 & 48.8 & 56.7 & 29.0 & 34.2 & 43.8 & 49.3 & 23.3 & 31.2 & 46.8 & 51.0 & 22.7 & 30.2 & 44.8 & 50.2\\
 \checkmark & \checkmark & \checkmark & \textbf{8.9} & \textbf{15.7} & \textbf{29.2} & \textbf{33.4} & \textbf{8.8} & \textbf{18.9} & \textbf{39.4} & \textbf{47.2} & \textbf{7.8} & \textbf{14.9} & \textbf{25.3} & \textbf{28.7} & \textbf{9.8} & \textbf{22.1} & \textbf{39.6} & \textbf{44.1} & \textbf{8.8} & \textbf{17.9} & \textbf{33.4} & \textbf{38.4}\\
 \hline
\end{tabular}
}
\caption{Influence of the DCT representation, the padding strategy, and the residual connections on 4 actions of H3.6M. Top: angle error; Bottom: 3D error (Models are trained on 3D). Note that, on average, all components of our model contribute to its accuracy.}
\vspace{-0.3cm}
\label{table-ablation-motion}
\end{table*}
\begin{table*}[ht]
\centering
\resizebox{\linewidth}{!}{%
\begin{tabular}{c|cccc|cccc|cccc|cccc|cccc}
  & \multicolumn{4}{c}{Walking} & \multicolumn{4}{c}{Eating} & \multicolumn{4}{c}{Smoking} & \multicolumn{4}{c}{Discussion}&\multicolumn{4}{c}{Average} \\
  & 80 & 160 & 320 & 400 & 80 & 160 & 320 & 400 & 80 & 160 & 320 & 400 & 80 & 160 & 320 & 400& 80 & 160 & 320 & 400 \\ \hline
  Fully-connected network & 0.20 & 0.34 & 0.54 & 0.61 & 0.18 & 0.31 & 0.53 & 0.66 & 0.22 & 0.43 & \textbf{0.85} & 0.83 & 0.28 & 0.64 & 0.87 & 0.93 & 0.22 & 0.43 & 0.70 & 0.76\\
 with pre-defined connectivity &  0.25 & 0.46 & 0.70 & 0.8 & 0.23 & 0.41 & 0.68 & 0.83 & 0.24 & 0.46 & 0.93 & 0.91 & 0.27 & 0.62 & 0.89 & 0.97 & 0.25 & 0.49 & 0.80 & 0.88\\
 with learnable connectivity & \textbf{0.18} & \textbf{0.31} & \textbf{0.49} & \textbf{0.56} & \textbf{0.16} & \textbf{0.29} & \textbf{0.50} & \textbf{0.62} & \textbf{0.22} & \textbf{0.41} & 0.86 & \textbf{0.80} & \textbf{0.20} & \textbf{0.51} & \textbf{0.77} & \textbf{0.85} & \textbf{0.19} & \textbf{0.38} & \textbf{0.66} & \textbf{0.71}\\
 \hline\hline
  Fully-connected network & 11.2 & 18.6 & 33.5 & 38.8 & 9.0 & 18.8 & \textbf{39.0} & 48.0 & 8.5 & 15.4 & 26.3 & 31.4 & 12.2 & 26.0 & 46.3 & 53.0 & 10.2 & 19.7 & 36.3 & 42.8\\
 with pre-defined connectivity & 25.6 & 44.6 & 80.3 & 96.8 & 16.3 & 31.9 & 62.4 & 78.8 & 11.6 & 21.4 & 34.6 & 38.6 & 20.7 & 38.7 & 62.5 & 69.9 & 18.5 & 34.1 & 59.9 & 71.0 \\
 with learnable connectivity & \textbf{8.9} & \textbf{15.7} & \textbf{29.2} & \textbf{33.4} & \textbf{8.8} & \textbf{18.9} & 39.4 & \textbf{47.2} & \textbf{7.8} & \textbf{14.9} & \textbf{25.3} & \textbf{28.7} & \textbf{9.8} & \textbf{22.1} & \textbf{39.6} & \textbf{44.1} & \textbf{8.8} & \textbf{17.9} & \textbf{33.4} & \textbf{38.4}\\
 \hline
\end{tabular}
}
\caption{Influence of GCNs and of learning the graph connectivity. Top: angle error; Bottom: 3D error. Note that GCNs with a pre-defined connectivity yield much higher errors than learning this connectivity as we do.}
\vspace{-0.3cm}
\label{table-ablation-fullyconnect}
\end{table*}
In Table~\ref{table-long-motion}, we report the long-term prediction errors on H3.6M in angle space and 3D space. In angle space, our approach yields the best results for 500ms, but a higher error than that of~\cite{gui2018adversarial} for 1000ms. Note that, based on our previous analysis, it is unclear if this is due to actual worse predictions or to the ambiguities of the angle representation. In terms of 3D errors, as shown in Table~\ref{table-long-motion}, our approach yields the best results by a large margin, particularly when trained using 3D coordinates.

\noindent\textbf{CMU-Mocap \& 3DPW.}
We report the results on the CMU dataset in terms of angle errors and 3D errors in Table~\ref{table-short-motion-cmu}, and those on the 3DPW in Table~\ref{table-short-long-3dpw}. In essence, the conclusions remain unchanged: Our method consistently outperforms the baselines for both short-term and long-term prediction, with the best results obtained when working directly with the 3D representation.

\subsection{Ablation Study}
To provide a deeper understanding of our approach, we now evaluate the influence of its several components. In particular, we investigate the importance of relying on the DCT to represent the temporal information. To this end, we compare our approach with a graph convolutional network trained using the joint angles or 3D coordinates directly as input. Furthermore, we study the influence of padding the input sequence with replicates of the last observed time step, instead of simply taking a shorter sequence as input, and the impact of using residual connections in our network. 

The results of these different experiments are provided in Table~\ref{table-ablation-motion}.
These results show that using our padding strategy provides a significant boost in accuracy, and so do the residual connections. In angle space, the influence of the DCT representation is sometimes small, but it remains important for some activities, such as "Discussion". By contrast, in 3D space, using the DCT representation yields significantly better results in all cases.

Finally, we evaluate the importance of using GCNs vs fully-connected networks and of learning the connectivity in the GCN instead of using a pre-defined adjacency matrix based on the kinematic tree. The results of these experiments, provided in Table~\ref{table-ablation-fullyconnect}, demonstrate the benefits of both using GCNs and learning the corresponding graph structure. Altogether, this ablation study evidences the importance of both aspects of our contribution: Using the DCT to model temporal information and learning the connectivity in GCNs to model spatial structure.

%% file: 5_conclusion.tex
\section{Conclusion}
In this paper, we have introduced an approach to human motion prediction that jointly encodes temporal information, via the use of the DCT, and spatial structure, via GCNs with learnable connectivity.~This leads to a compact, feed-forward network with proven highly effectiveness for the prediction task.  Our approach achieves state-of-the-art results on standard human motion prediction benchmarks.  Experiments have also revealed an interesting phenomenon: evaluating motion prediction in angle space is unreliable, as the angle representation has ambiguities such that two very different sets of angles can share the same 3D pose. We thus argue that, in contrast to the main trend in the literature, motion prediction should be performed in 3D space.~This was confirmed by our experiments, in which the models trained on 3D coordinates consistently outperform those trained on angles. 
Our future work will focus on a systematic analysis of this phenomenon.
\vspace{-0.1cm}
\section*{Acknowledgements}
\vspace{-0.1cm}
This research was supported in part by the Australia Centre for Robotic Vision (CE140100016), the Australia Research Council DECRA Fellowship (DE180100628), ARC Discovery Grant (DP190102261) and LIEF(LE190100080).  The authors would like to thank NVIDIA for the donated GPU (Titan V) and the GPU cluster in NCI Australia.

%% file: HumanDynamicsSupplementary.tex
\title{Learning Trajectory Dependencies for Human Motion Prediction \\
-----Supplementary Material-----}

\author{Wei Mao$^1$, \;\;Miaomiao Liu$^{1,3}$,\;\; Mathieu Salzmann$^2$,\;\; Hongdong Li$^{1,3}$\\
$^1$Australian National University, $^2$CVLab, EPFL,$^3$Australia Centre for Robotic Vision\\
{\tt\small \{wei.mao, miaomiao.liu, hongdong.li\}@anu.edu.au,}\;\;{\tt\small mathieu.salzmann@epfl.ch}
}

\maketitle
\thispagestyle{empty}

\section{Datasets}
Below, we provide more detail on the datasets used in our experiments.

\vspace{1mm}
\noindent\textbf{Human3.6M.} In H3.6M, each pose has 32 joints. 
Removing the global rotation, translation and constant angles, leaves us with a 48 dimensional vector for each human motion, denoting the exponential map representation of the joint angles. Furthermore, a 3D human pose can also be represented by a 66 dimensional vector of 3D coordinates after removing the global rotation, translation and stationary joints across time.
We use the same training and test split as previous work~\cite{Martinez_2017_CVPR,LiZLL18,gui2018adversarial}. That is, we test our model on the same image sequence of subject 5 as previous work~\cite{Martinez_2017_CVPR,LiZLL18,gui2018adversarial}.
For training, we keep subject 11 as validation set to choose the best model (the one that achieves the least average error across all future frames) and use the remaining 5 subjects as training set. 

\vspace{1mm}
\noindent\textbf{CMU-Mocap.} 
In this dataset, we use a 64 dimensional vector to represent every human pose by removing the global rotation, translation and joint angles with constant values. Each component of the vector denotes the exponential map representation of the joint angle. We further use 75 dimensional vectors for the 3D joint coordinate representation. We do not use a validation set due to limited training data.

\vspace{1mm}
\noindent\textbf{3DPW.} The human skeleton in this dataset uses 24 joints, yielding a 72 dimensional vector for the angle representation. For the 3D joint coordinate one, we obtain a 69 dimensional vector after removing the global translation.

\section{Visualizing the Results on H3.6M in Video}
We provide more visualization of the results on H3.6M in a video (See the supplementary video). In particular, the video compares our approach with the state of the art on periodic actions, such as walking, and aperiodic ones, such as eating and direction. Our approach shows better performance than the state-of-the-art ones.

Furthermore, in the video, we provide additional (quantitative and qualitative) visualization of cases where large errors are observed according to the angle representation but small errors in 3D space. This confirms that ambiguities exist in angle space for human motion prediction.

\section{Visualizing the Results on CMU-Mocap}
\begin{figure}[!t]
    \centering
    \includegraphics[width=\linewidth]{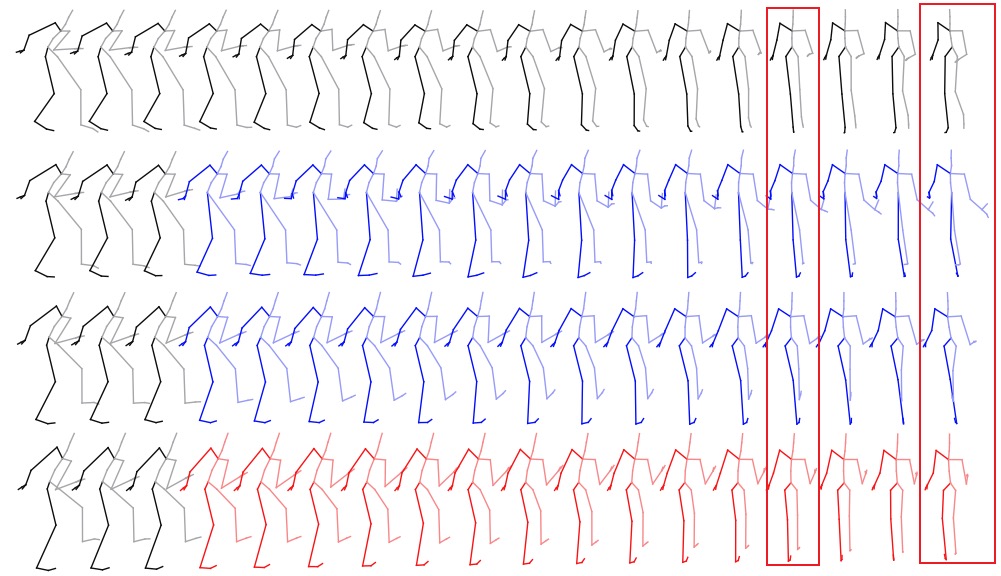}
    \caption{Motion prediction in 3D space on the ``basketball action of CMU-Mocap. From top to bottom: Ground truth, results of \cite{Martinez_2017_CVPR}, results of \cite{LiZLL18} and our results.
    The highlighted results in the box show that we can make better predictions on the legs and arms of the subject.
    }
    \label{fig:jumping_6}
\end{figure}

\begin{figure}[!t]
    \centering
    \includegraphics[width=\linewidth]{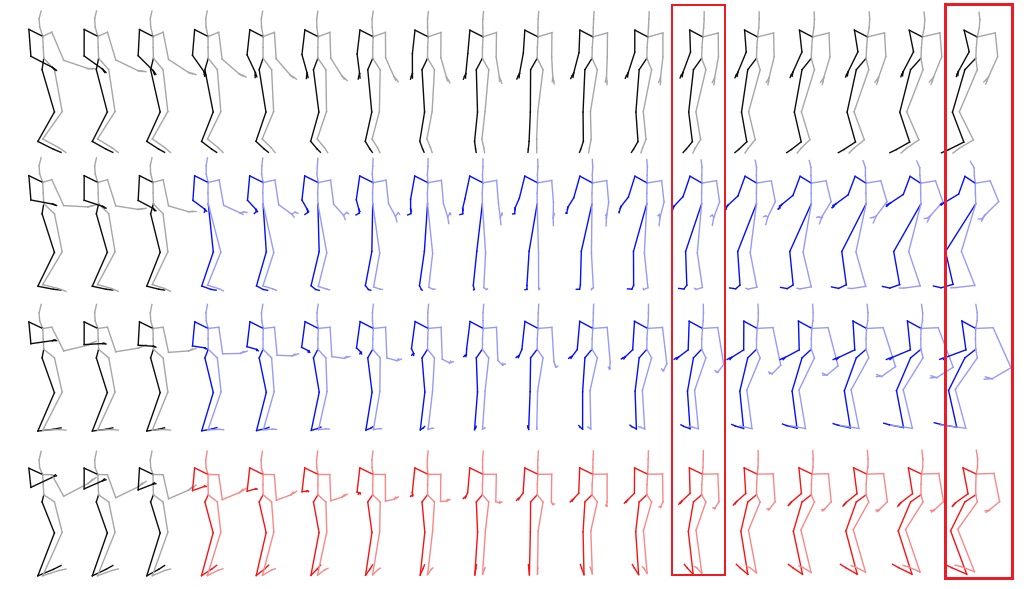}
    \caption{Motion prediction in 3D space on the ``basketball signal" action of CMU-Mocap. From top to bottom: Ground truth, results of \cite{Martinez_2017_CVPR}, results of \cite{LiZLL18} and our results.}
    \label{fig:basketball_signal_3}
\end{figure}

\begin{figure}[!t]
    \centering
    \includegraphics[width=\linewidth]{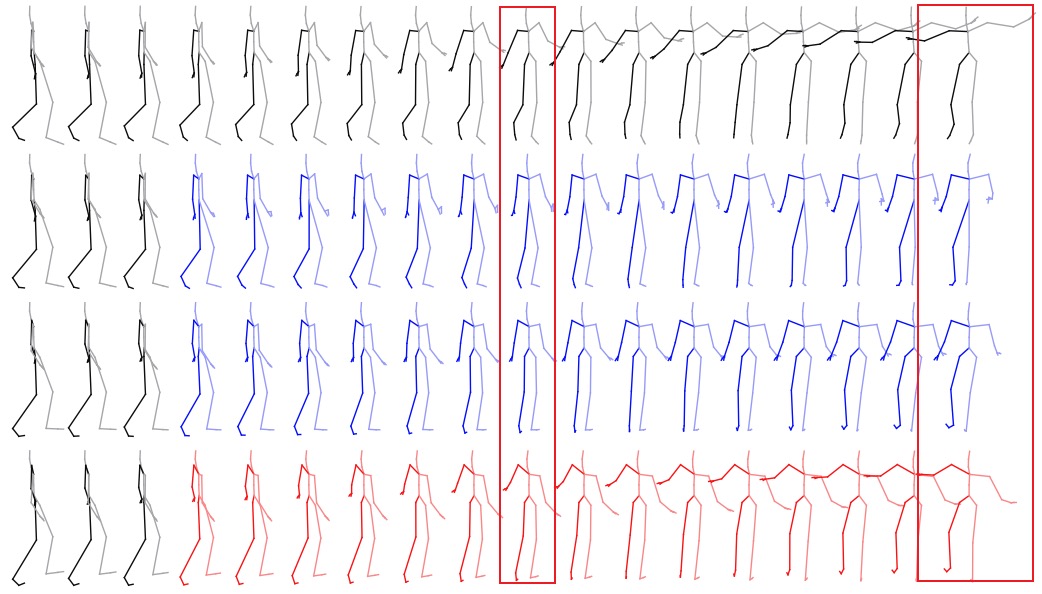}
    \caption{Motion prediction in 3D space on the ``directing traffic" action of CMU-Mocap. From top to bottom: Ground truth, results of \cite{Martinez_2017_CVPR}, results of \cite{LiZLL18} and our results.}
    \label{fig:directing_traffic_7}
\end{figure}

We provide a qualitative visualisation of the 3D human pose prediction on the ``basketball", ``basketball signal" and ``direction traffic" actions of the CMU-Mocap dataset in Fig.~\ref{fig:jumping_6}, Fig.~\ref{fig:basketball_signal_3} and Fig.~\ref{fig:directing_traffic_7}, respectively. Again, our approach outperforms the state-of-the-art ones (see highlighted poses).


\section{Number of DCT Coefficients}
In this section, we first present the intuition behind using fewer DCT coefficients to represent the whole sequence. We then compare the performance of using different number of DCT coefficients.
\subsection{Using Fewer Coefficients}
\begin{figure}[!t]
    \centering
    \includegraphics[width=\linewidth]{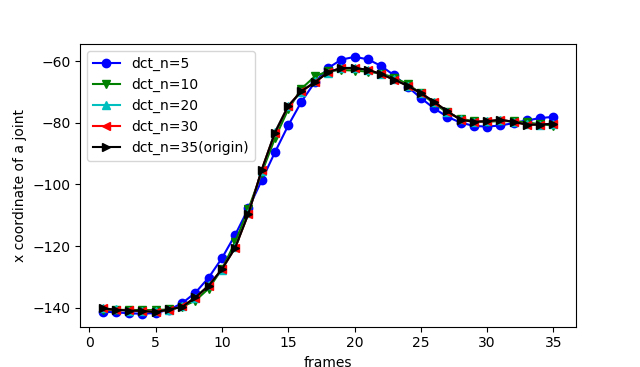}
    \caption{Temporal trajectory  of the $x$ coordinate of one joint reconstructed using different number of DCT coefficients.}
    \label{fig:dct_n}
\end{figure}

Given a smooth trajectory, it is possible to discard some high frequency DCT coefficients without losing prediction accuracy. To evidence this, in~Fig.~\ref{fig:dct_n}, we show the effect of the number of DCT components in reconstructing a sequence of 35 frames for the one human joint predicted using our approach. 
Note that, since we use 35 frames, 35 DCT coefficients yield a lossless reconstruction. Nevertheless, even 10 DCT coefficients are enough to reconstruct the trajectory with a very low error. This is due to the smoothness of the joint trajectory in 3D space.  
\begin{figure*}[!t]
    \centering
    \includegraphics[width=\linewidth]{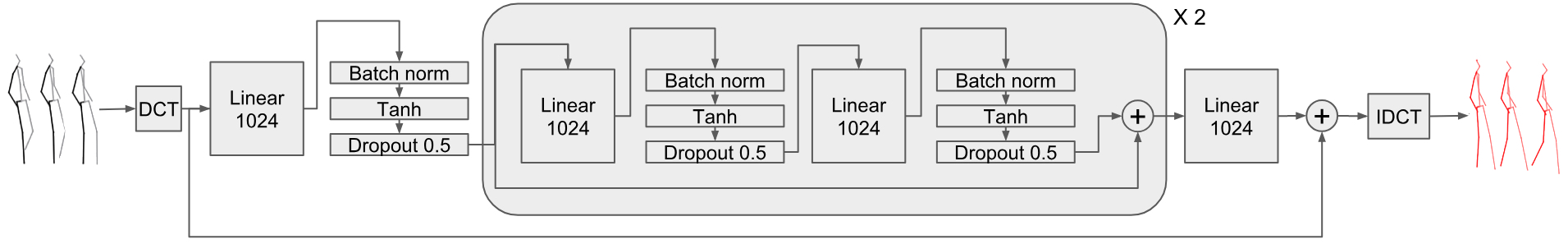}
    \caption{Fully-connected Network Structure}
    \label{fig:fully-connected-net}
\end{figure*}

\comment{
During our experiments, we found that the trajectory of 3D coordinate is always smooth and we can apply this compression on DCT coefficients while the trajectory of exponential map representation of joint rotation is not always smooth. So that in our experiment in the paper, we use less DCT coefficients when predicting 3D pose and use full DCT coefficients when predicting angles. We will show the comparison of using different number of DCT coefficients.
}
\subsection{Results on H3.6M}
\begin{figure}[t]
    \centering
    \includegraphics[width=\linewidth]{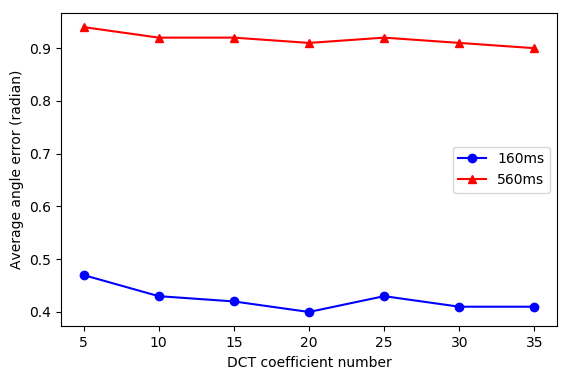}
    \caption{Average angle prediction error over 4 actions ("walking","eating","smoking","discussion") using different number of DCT coefficients at 160ms (\textcolor{blue}{blue}) and 560ms (\textcolor{red}{red}).}
    \label{fig:dct-n-angle}
\end{figure}
\begin{figure}[ht]
    \centering
    \includegraphics[width=\linewidth]{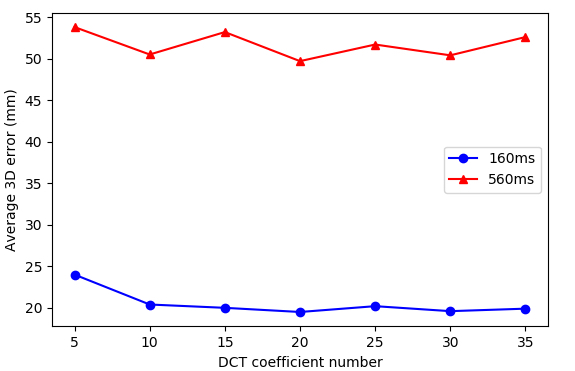}
    \caption{Average 3D prediction error over 4 actions (``walking",``eating",``smoking",``discussion") using different number of DCT coefficients at 160ms (\textcolor{blue}{blue}) and 560ms (\textcolor{red}{red}).}
    \label{fig:dct-n-3d}
\end{figure}
\textbf{Experiment setup.} 
Based on the previous discussion, we perform more experiments to evaluate the influence of the number of input DCT coefficients on human motion prediction. In the following experiments, we assume that we observe 10 frames to predict the future 25 frames. Following the same formulation as in our submission, the observed sequence is padded with the last observed frame replicated 25 times and then transformed to DCT coefficients. The target is the DCT coefficients of the whole sequence (35 frames).
We perform several experiments by preserving different number of DCT coefficients. For instance, `dct\_n=5' means that we only use the first 5 DCT coefficients for temporal reconstruction. The experiments are performed on both 3D and angle representation.


Fig.~\ref{fig:dct-n-angle} shows the error for short-term prediction at $160$ms and long-term prediction at $560$ms in angle representation as a function of the number of DCT coefficients. In general, the angle error decreases with the increase of number of DCT coefficients. Similarly, in Fig.~\ref{fig:dct-n-3d},
we plot the motion prediction error in 3D coordinates at $160$ms and $560$ms as a function of the number of DCT coefficients. Here, $10$ DCT coefficients already give a very small prediction error. Interestingly, when we use more DCT coefficients, the average error sometimes increases (see the plot for prediction at 560ms). This pattern confirms our argument in the submission that the use of truncated DCT coefficients can prevent a model from generating jittery motion, because the
3D coordinate representation of human motion trajectory is smooth.

\begin{figure}[t]
    \centering
    \includegraphics[width=\linewidth]{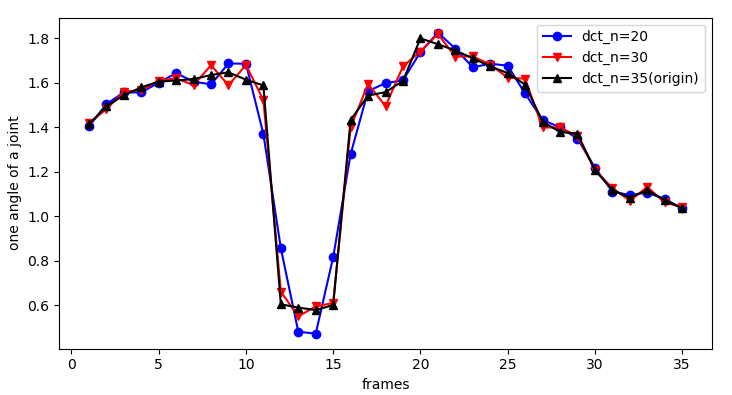}
    \caption{ The temporal trajectory of one joint angle reconstructed using different number of DCT coefficients. Note that the trajectory is not smooth and has large jumps. The full frequency (35 DCT coefficients) leads to lossless temporal reconstruction of the trajectory.  
    }
    \label{fig:dct-n-angle-trajectory}
\end{figure}

To analyse the different patterns of the prediction error w.r.t. the number of DCT coefficients shown in angle representation (Fig.~\ref{fig:dct-n-angle}) and 3D representation (Fig.~\ref{fig:dct-n-3d}), we looked into the dataset and found that there are large discontinuities in the trajectories of angles. As shown in Fig.~\ref{fig:dct-n-angle-trajectory}, these large jumps make the reconstruction of trajectories with fewer DCT coefficients lossy.

In summary, we can discard some of the high frequency coefficients to achieve better performance in 3D space. In our experiments, we use the first 15 DCT coefficients as input to our network for short-term prediction and 30 coefficients for long-term prediction in 3D space.
As the joint trajectory in angle representation is not smooth and has large discontinuities, we therefore take the full frequency as input to our network for motion prediction in angle representation.  In our experiments, we therefore use 20 DCT coefficients as input to our network for short-term prediction and 35 for long-term prediction in angle representation.

\section{Ablation Study Details}
\textbf{Fully-connected Network.}
In our ablation study, we also compare the motion prediction using a graph network with that of a fully-connected network structure. We apply the same process of encoding temporal information via the DCT. Before being fed to the network, the DCT coefficients of the past sequence padded with last frame are flattened to a vector and the network learns the residual between the past temporal encoding and the future one. To this end, we adopt the network structure shown in Fig.~\ref{fig:fully-connected-net}. Instead of using graph convolutional layers, we rely on 2 fully connected layers with residual connections. We additionally use two fully connected layers at the start of the network for encoding the DCT coefficients and at the end for decoding the feature to the residual of the DCT coefficient. 

The implementation details for this network are the same as our Graph Convolutional Network. We implemented this network using Pytorch~\cite{paszke2017automatic}, and we used ADAM~\cite{kingma2014adam} to train this model.
The learning rate was set to 0.0005 with a 0.96 decay every two epochs. The batch size was set to 16 and the gradients were clipped to a maximum $\ell$2-norm of 1. The model was trained for 50 epochs. As reported in the submission, the fully-connected network structure cannot learn a better representation than the Graph Convolutional Network.
\section{Mean Pose Problem}
\begin{figure}[ht]
\begin{center}
   \includegraphics[width=0.97\linewidth]{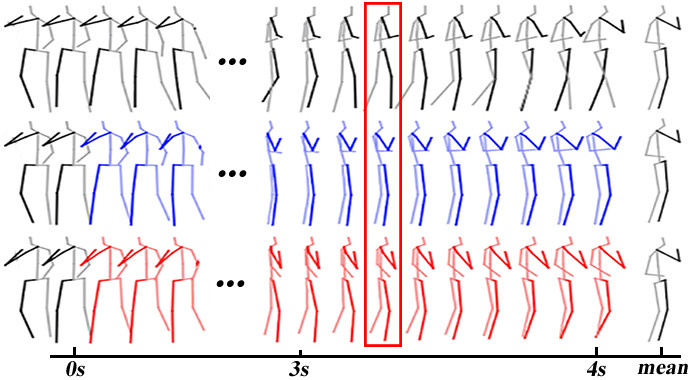}
\end{center}
\vspace{-0.3cm}
   \caption{Prediction up to 4 seconds for the Phoning action of Human3.6m. From top to bottom, we show the ground truth, the poses predicted by~\cite{Martinez_2017_CVPR} 
   , and by our method. 
   Note that, after the highlighted frame, the poses predicted by the RNN of~\cite{Martinez_2017_CVPR} have indeed converged to the mean pose (shown in the last column), whereas in our predictions the legs continue to move.}
\label{fig:mean-pose}
\end{figure}

 As explained in~\cite{LiZLL18}, the mean pose problem typically occurs when using recurrent neural networks (RNNs) to encode temporal dynamics, where the past information may vanish during long propagation paths. By not relying on RNNs, but directly encoding the trajectory of the whole sequence, our method inherently prevents us from losing the past information.~This is evidenced by Fig.~\ref{fig:mean-pose}, where our method yields poses significantly further from the mean pose than the RNN-based method~\cite{Martinez_2017_CVPR}.

%% file: HumanDynamicsModelling.bbl
\begin{thebibliography}{10}\itemsep=-1pt

\bibitem{akhter2009nonrigid}
Ijaz Akhter, Yaser Sheikh, Sohaib Khan, and Takeo Kanade.
\newblock Nonrigid structure from motion in trajectory space.
\newblock In {\em Advances in neural information processing systems}, pages
  41--48, 2009.

\bibitem{ArjovskyB17}
Mart{\'{\i}}n Arjovsky and L{\'{e}}on Bottou.
\newblock Towards principled methods for training generative adversarial
  networks.
\newblock In {\em ICLR}, 2017.

\bibitem{brand2000style}
Matthew Brand and Aaron Hertzmann.
\newblock Style machines.
\newblock In {\em Proceedings of the 27th annual conference on Computer
  graphics and interactive techniques}, pages 183--192. ACM
  Press/Addison-Wesley Publishing Co., 2000.

\bibitem{BrunaZSL13}
Joan Bruna, Wojciech Zaremba, Arthur Szlam, and Yann LeCun.
\newblock Spectral networks and locally connected networks on graphs.
\newblock In {\em ICLR}, 2014.

\bibitem{Butepage_2017_CVPR}
Judith Butepage, Michael~J. Black, Danica Kragic, and Hedvig Kjellstrom.
\newblock Deep representation learning for human motion prediction and
  classification.
\newblock In {\em CVPR}, July 2017.

\bibitem{defferrard2016convolutional}
Micha{\"e}l Defferrard, Xavier Bresson, and Pierre Vandergheynst.
\newblock Convolutional neural networks on graphs with fast localized spectral
  filtering.
\newblock In {\em Advances in neural information processing systems}, pages
  3844--3852, 2016.

\bibitem{fragkiadaki2015recurrent}
Katerina Fragkiadaki, Sergey Levine, Panna Felsen, and Jitendra Malik.
\newblock Recurrent network models for human dynamics.
\newblock In {\em ICCV}, pages 4346--4354, 2015.

\bibitem{gong2011multi}
Haifeng Gong, Jack Sim, Maxim Likhachev, and Jianbo Shi.
\newblock Multi-hypothesis motion planning for visual object tracking.
\newblock In {\em ICCV}, pages 619--626. IEEE, 2011.

\bibitem{gui2018adversarial}
Liang-Yan Gui, Yu-Xiong Wang, Xiaodan Liang, and Jos{\'e}~MF Moura.
\newblock Adversarial geometry-aware human motion prediction.
\newblock In {\em ECCV}, pages 786--803, 2018.

\bibitem{h36m_pami}
Catalin Ionescu, Dragos Papava, Vlad Olaru, and Cristian Sminchisescu.
\newblock Human3.6m: Large scale datasets and predictive methods for 3d human
  sensing in natural environments.
\newblock {\em IEEE Transactions on Pattern Analysis and Machine Intelligence},
  36(7):1325--1339, jul 2014.

\bibitem{JainZSS16}
Ashesh Jain, Amir~Roshan Zamir, Silvio Savarese, and Ashutosh Saxena.
\newblock Structural-rnn: Deep learning on spatio-temporal graphs.
\newblock In {\em CVPR}, pages 5308--5317, 2016.

\bibitem{kingma2014adam}
Diederik~P Kingma and Jimmy Ba.
\newblock Adam: A method for stochastic optimization.
\newblock {\em arXiv preprint arXiv:1412.6980}, 2014.

\bibitem{kipf2016semi}
Thomas~N Kipf and Max Welling.
\newblock Semi-supervised classification with graph convolutional networks.
\newblock In {\em ICLR}, 2017.

\bibitem{kiros2015skip}
Ryan Kiros, Yukun Zhu, Ruslan~R Salakhutdinov, Richard Zemel, Raquel Urtasun,
  Antonio Torralba, and Sanja Fidler.
\newblock Skip-thought vectors.
\newblock In {\em Advances in neural information processing systems}, pages
  3294--3302, 2015.

\bibitem{koppula2013anticipating}
Hema~Swetha Koppula and Ashutosh Saxena.
\newblock Anticipating human activities for reactive robotic response.
\newblock In {\em IROS}, page 2071. Tokyo, 2013.

\bibitem{LiZLL18}
Chen Li, Zhen Zhang, Wee~Sun Lee, and Gim~Hee Lee.
\newblock Convolutional sequence to sequence model for human dynamics.
\newblock In {\em CVPR}, pages 5226--5234, 2018.

\bibitem{Martinez_2017_CVPR}
Julieta Martinez, Michael~J. Black, and Javier Romero.
\newblock On human motion prediction using recurrent neural networks.
\newblock In {\em CVPR}, July 2017.

\bibitem{paden2016survey}
Brian Paden, Michal {\v{C}}{\'a}p, Sze~Zheng Yong, Dmitry Yershov, and Emilio
  Frazzoli.
\newblock A survey of motion planning and control techniques for self-driving
  urban vehicles.
\newblock {\em IEEE Transactions on intelligent vehicles}, 1(1):33--55, 2016.

\bibitem{pascanu2013difficulty}
Razvan Pascanu, Tomas Mikolov, and Yoshua Bengio.
\newblock On the difficulty of training recurrent neural networks.
\newblock In {\em ICML}, pages 1310--1318, 2013.

\bibitem{paszke2017automatic}
Adam Paszke, Sam Gross, Soumith Chintala, Gregory Chanan, Edward Yang, Zachary
  DeVito, Zeming Lin, Alban Desmaison, Luca Antiga, and Adam Lerer.
\newblock Automatic differentiation in pytorch.
\newblock In {\em NIPS-W}, 2017.

\bibitem{sutskever2011generating}
Ilya Sutskever, James Martens, and Geoffrey~E Hinton.
\newblock Generating text with recurrent neural networks.
\newblock In {\em ICML}, pages 1017--1024, 2011.

\bibitem{velivckovic2017graph}
Petar Veli{\v{c}}kovi{\'c}, Guillem Cucurull, Arantxa Casanova, Adriana Romero,
  Pietro Lio, and Yoshua Bengio.
\newblock Graph attention networks.
\newblock In {\em ICLR}, 2018.

\bibitem{vonMarcard2018}
Timo von Marcard, Roberto Henschel, Michael Black, Bodo Rosenhahn, and Gerard
  Pons-Moll.
\newblock Recovering accurate 3d human pose in the wild using imus and a moving
  camera.
\newblock In {\em ECCV}, 2018.

\bibitem{wang2008gaussian}
Jack~M Wang, David~J Fleet, and Aaron Hertzmann.
\newblock Gaussian process dynamical models for human motion.
\newblock {\em IEEE transactions on pattern analysis and machine intelligence},
  30(2):283--298, 2008.

\bibitem{yan2018spatial}
Sijie Yan, Yuanjun Xiong, and Dahua Lin.
\newblock Spatial temporal graph convolutional networks for skeleton-based
  action recognition.
\newblock In {\em Thirty-Second AAAI Conference on Artificial Intelligence},
  2018.

\end{thebibliography}


\begin{thebibliography}{1}\itemsep=-1pt

\bibitem{gui2018adversarial}
Liang-Yan Gui, Yu-Xiong Wang, Xiaodan Liang, and Jos{\'e}~MF Moura.
\newblock Adversarial geometry-aware human motion prediction.
\newblock In {\em ECCV}, pages 786--803, 2018.

\bibitem{kingma2014adam}
Diederik~P Kingma and Jimmy Ba.
\newblock Adam: A method for stochastic optimization.
\newblock {\em arXiv preprint arXiv:1412.6980}, 2014.

\bibitem{LiZLL18}
Chen Li, Zhen Zhang, Wee~Sun Lee, and Gim~Hee Lee.
\newblock Convolutional sequence to sequence model for human dynamics.
\newblock In {\em CVPR}, pages 5226--5234, 2018.

\bibitem{Martinez_2017_CVPR}
Julieta Martinez, Michael~J. Black, and Javier Romero.
\newblock On human motion prediction using recurrent neural networks.
\newblock In {\em CVPR}, July 2017.

\bibitem{paszke2017automatic}
Adam Paszke, Sam Gross, Soumith Chintala, Gregory Chanan, Edward Yang, Zachary
  DeVito, Zeming Lin, Alban Desmaison, Luca Antiga, and Adam Lerer.
\newblock Automatic differentiation in pytorch.
\newblock In {\em NIPS-W}, 2017.

\end{thebibliography}
